\documentclass[comsoc,journal]{IEEEtran} 

\usepackage{pdflscape}  
\usepackage{booktabs}   
\usepackage{adjustbox}  
\usepackage{array}      
\usepackage{siunitx}    
\usepackage{tablefootnote}
\usepackage{threeparttable}
\usepackage{float}
\usepackage{cite}
\usepackage{setspace}
\usepackage{graphicx}
\usepackage{newtxmath}
\usepackage{hyperref}
\usepackage{amsfonts}
\usepackage{makecell}
\usepackage{amsmath}
\usepackage{algorithm,algcompatible}
\usepackage{fancyhdr}
\usepackage{xcolor}
\usepackage{multirow}
\usepackage{flafter}
\usepackage{url}
\usepackage{bm} 
\usepackage{enumerate}
\usepackage{subfigure} 
\usepackage{epstopdf}
\usepackage{booktabs}
\usepackage{multirow}
\providecommand{\keywords}[1]{\textbf{\textit{Index terms---}} #1}

\begin{document}
\title{Improve the Training Efficiency of DRL for Wireless Communication Resource Allocation: The Role of Generative Diffusion Models}
%
\author{Xinren~Zhang,
        Jiadong~Yu,\IEEEmembership{ Member,~IEEE} 
\thanks{
X. Zhang and J. Yu are with the Internet of Things Thrust at the Hong Kong University of Science and Technology (Guangzhou).
}}
\maketitle

\begin{abstract}
Dynamic resource allocation in mobile wireless networks involves complex, time-varying optimization problems, motivating the adoption of deep reinforcement learning (DRL). However, most existing works rely on pre-trained policies, overlooking dynamic environmental changes that rapidly invalidate the policies. Periodic retraining becomes inevitable but incurs prohibitive computational costs and energy consumption—critical concerns for resource-constrained wireless systems. We identify three root causes of inefficient retraining: high-dimensional state spaces, suboptimal action spaces exploration-exploitation trade-offs, and reward design limitations.
To overcome these limitations, we propose Diffusion-based Deep Reinforcement Learning (D2RL), which leverages generative diffusion models (GDMs) to holistically enhance all three DRL components. GDMs’ iterative refinement process and distribution modelling enable (1) the generation of diverse state samples to improve environmental understanding, (2) balanced action space exploration to escape local optima, and (3) the design of discriminative reward functions that better evaluate action quality. Our framework operates in two modes: Mode I leverages GDMs to explore reward spaces and design discriminative reward functions that rigorously evaluate action quality, while Mode II synthesizes diverse state samples to enhance environmental understanding and generalization. Extensive experiments demonstrate that D2RL achieves faster convergence and reduced computational costs over conventional DRL methods for resource allocation in wireless communications while maintaining competitive policy performance. This work underscores the transformative potential of GDMs in overcoming fundamental DRL training bottlenecks for wireless networks, paving the way for practical, real-time deployments.

\end{abstract}


\section{Introduction}
\label{sec:intro}

Mobile wireless networks face challenges in resource allocation due to fluctuating channel conditions and multi-objective optimization \cite{hieu2021optimal,gao2024uav}. Deep reinforcement learning (DRL) has emerged as a promising solution for these complex resource allocation problems, offering the ability to learn optimal policies through interaction with the environment without requiring explicit mathematical modeling of system dynamics\cite{he2023full,chauhan2024full,nayak2024drl}.

While pre-trained DRL models achieve static optimization\cite{kumar2020conservative, agarwal2020optimistic}, their inefficiency in dynamic environments necessitates periodic retraining \cite{muhammad2021deep}. In such cases, training efficiency, characterized by both the convergence speed toward optimal policies and the computational resources consumed during the training process, becomes crucial\cite{jin2022vwp}. The retraining process consumes substantial computational resources, and inefficient training can lead to excessive energy consumption and resource utilization. Therefore, when considering the application of DRL in mobile wireless environments, training efficiency is an essential factor that requires careful consideration.

Three key elements of DRL - state space, action space, and reward - fundamentally influence the training efficiency. Each of these elements presents unique challenges that can significantly impact the gradient updates and, consequently, the training efficiency. State data must capture relevant environmental information while remaining computationally manageable, yet comprehensive state data acquisition is often resource-intensive and time-consuming \cite{qian2024offline,wang2021provably}. Limited or poor-quality state data can lead to incomplete environmental understanding and extended convergence times \cite{wang2022vrl3}. Moreover, insufficient state diversity can induce overfitting, where agents become overly specialized in specific scenarios, compromising their generalization capabilities in real-world deployments\cite{mohi2024optimizing}. Action selection requires balancing exploration and exploitation, where insufficient exploration can result in convergence to suboptimal policies\cite{chen2023structure}. Reward design must provide meaningful feedback for learning, but traditional simple reward functions\cite{michaud2020understanding,lyu2020movement} often prove inadequate for complex wireless communication scenarios with multiple competing objectives. While the integration of expert knowledge into reward design represents a potential solution\cite{du2024enhancing}, offering more informed feedback and helping avoid suboptimal strategies, it introduces its own limitations. Excessive reliance on expert-guided rewards can constrain the agent's exploration space, potentially leading to convergence toward local optima rather than discovering innovative or globally optimal solutions.

Recent advances in generative diffusion models (GDMs) \cite{sohl2015deep} have demonstrated remarkable capabilities in generating diverse and high-quality samples through denoising processes. This aligns well with the key requirements in addressing DRL challenges: the need for diverse state representation, effective action space exploration, and comprehensive reward signals. The iterative refinement process in GDMs naturally supports exploration by gradually transforming noise into meaningful samples, while their ability to model complex distributions enables the generation of diverse and representative samples. Leveraging these properties, \cite{du2024enhancing} has successfully applied GDMs to enhance action space exploration in DRL, demonstrating improved training efficiency and convergence. This success suggests promising potential for extending GDM applications to address the challenges in state representation and reward design as well.

\textcolor{black}{To address the challenges of DRL training efficiency in full-duplex (FD) wireless communications, }this paper presents a systematic investigation into the integration of GDMs with the three fundamental elements of DRL. While previous research has primarily focused on action space exploration using GDMs, we propose a more comprehensive framework called Diffusion-based Deep Reinforcement Learning (D2RL) that leverages multiple GDM operational modes to address the complex challenge of resource allocation in dynamic, multi-objective environments. By facilitating thorough exploration across state, action, and reward spaces, D2RL significantly enhances training efficiency, achieving faster convergence with reduced computational overhead. These improvements mark a substantial step toward making DRL more viable for practical deployment in real-world wireless communication systems.
\vspace{-8pt}
\subsection{Contributions}
Our contributions are summarized as follows:
\begin{itemize}
\item We present a generalized mobile wireless communication system model. To address the resource allocation problem of the model using DRL, we analyze the challenges associated with three key elements of DRL in mobile wireless communications: state space, action space, and reward space. Through the examination of gradient updates during the learning process, we provide theoretical analysis that quantifies the impact of these three spaces on DRL training efficiency.
\item  We propose D2RL framework utilizing GDMs in three key elements: generating diverse state samples for improved environmental understanding, exploring action spaces for better policy learning, and enhancing reward function design for more informative feedback. We conduct a systematic study of their individual and collaborative effects on training efficiency.
\item We propose two distinct modes of GDMs for D2RL framework based on the availability of original dataset. In Mode I, GDMs are employed to explore reward spaces, creating more informative and discriminative reward functions that better capture the quality of agent-generated actions. In Mode II, GDMs generate diverse and representative state samples to expand the agent's environmental understanding and improve generalization to unseen situations.
\item We conduct extensive experiments in an FD wireless communication system. The results demonstrate improvements in both training efficiency and final performance metrics. Specifically, our approach achieves faster convergence and requires fewer computational resources compared to conventional DRL methods, while maintaining the quality of learned policies.
\end{itemize}
\vspace{-8pt}
\subsection{Related Works}

\subsubsection{Deep Reinforcement Learning for Resource Allocation in Wireless Communications}

Resource allocation in wireless networks has attracted significant research attention, particularly through the application of DRL approaches. Several works have proposed innovative frameworks to address various challenges in different network scenarios. In heterogeneous cellular networks, \cite{zhi2022deep} proposed a distributed multi-agent deep Q-network algorithm for joint mode selection and channel allocation in D2D-enabled networks, maximizing system sum-rate through partial information sharing. For UAV-assisted networks, \cite{luong2021deep} combined deep Q-learning with difference of convex algorithm (DCA) to jointly optimize UAV positions and resource allocation. In addition, \cite{wang2023hybrid} developed a hybrid hierarchical DRL framework integrating double deep Q networks (D3QN) and soft Actor-Critic (SAC) for secure transmission in intelligent reflective surface (IRS)-assisted networks. Moreover,
several frameworks have been proposed for energy-efficient resource allocation. \cite{wang2020drl} introduced three complementary approaches - the discrete DRL-based resource allocation (DDRA), continuous DRL-based resource allocation (CDRA), and joint DRL and optimization resource allocation (DORA) - incorporating event-triggered learning to balance complexity and performance. In semantic communications, \cite{zhang2023drl} developed a deep deterministic policy gradient (DDPG)-based framework optimizing bandwidth and semantic compression ratio. For ultra-reliable and low-latency communications (URLLC) systems, \cite{tran2023multi} proposed a multi-agent (MA) DRL framework combining MA dueling double deep Q network (MA3DQN), MA double deep Q network (MA2DQN), and  MA deep Q network (MADQN) algorithms to ensure ultra-reliable and low-latency requirements while optimizing resource allocation.

While these existing approaches have made significant contributions to resource allocation in wireless networks, they primarily focus on achieving optimal performance metrics without explicitly considering the training efficiency. This paper aims to explore algorithms that can further accelerate the training process of utilizing DRL to solve resource allocation problems in wireless communications. Despite the fact that previous research has achieved satisfactory training results, the question remains whether the learning speed can be further expedited to save computational resources and improve efficiency. To address this issue, this paper delves into three critical components of DRL: action, state, and reward spaces. Through the exploration of these spaces, we seek to uncover previously overlooked information that could potentially speed up the training process.

\subsubsection{Generative Diffusion Models for Deep Reinforcement Learning}
Diffusion models have emerged as a powerful paradigm in various domains since their introduction by \cite{sohl2015deep}, who proposed a deep unsupervised learning framework that learns to reverse a Markov diffusion process for efficient learning and sampling. Building upon this foundation, several works have extended diffusion models to decision-making and planning tasks. For instance, \cite{janner2022planning} introduced Diffuser, a diffusion probabilistic model for trajectory optimization that enables flexible long-horizon behavior synthesis through iterative denoising. In the context of vehicular metaverses, \cite{kang2024hybrid} developed a hybrid-generative diffusion model that uniquely handles both discrete and continuous actions for vehicle twin migration, demonstrating superior convergence and efficiency. Recent work has further expanded the application of diffusion models to edge computing and AI-generated content. \cite{du2024diffusion} proposed deep diffusion soft actor-critic (D2SAC), integrating diffusion-based AI-Generated Optimal Decision (AGOD) algorithm with soft actor-critic architecture for edge-enabled artificial intelligence-generated content (AIGC) service provider selection. Most recently, \cite{qin2024diffusiongpt} introduced DiffusionGPT, a unified framework that leverages large language models to integrate diverse prompt types with domain-expert models through a Tree-of-Thought structure, advancing the capabilities of text-to-image generation systems.

While these existing works have demonstrated the effectiveness of diffusion models across various domains, they have primarily focused on applying GDMs to specific aspects of decision-making tasks rather than exploring their potential for comprehensive DRL enhancement. This paper aims to accelerate the DRL training process in mobile wireless communication through GDM applications. Since prior research has demonstrated powerful exploration capabilities of GDMs in expanding action spaces within DRL frameworks, our work extends their application to two additional critical components of the DRL paradigm. We leverage GDMs to design more sophisticated reward functions capable of providing higher-quality feedback for action evaluation. Moreover, we utilize GDMs to augment existing state space datasets, thereby reducing the resource requirements for environmental interactions. In addition, we compared the impact of collaboratively using GDM on the learning speed of DRL in exploring these three spaces.

The remainder of this paper is organized as follows: Section \ref{sec:Preliminaries} presents the preliminaries, while Section \ref{sec:system} formulates the system model and problem. 
In Section \ref{sec:algo}, we propose our framework: D2RL, designed to address the problem.
The performance evaluation of D2RL is presented in Section \ref{sec:evaluation}, discussing the results and insights gained from the experiments. Finally, Section \ref{sec:conclusion} concludes our contribution and findings.

\vspace{-6pt}
\section{Preliminaries}\label{sec:Preliminaries}
In this section, we introduce the foundational concepts and relevant background, focusing on DRL in resource allocation for wireless communication systems and an overview of GDMs.
\vspace{-8pt}
\subsection{Deep Reinforcement Learning in Resource Allocation for Wireless Communication}
In DRL, the decision-making process of an agent is formalized as a Markov Decision Process (MDP), defined by the tuple $(\mathcal{S}, \mathcal{A}, {R}, \gamma)$, where $\mathcal{S}$ is the observation space, $\mathcal{A}$ is the action space, ${R}:\ \mathcal{S} \times \mathcal{A} \rightarrow \mathbb{R}$ is the reward function, and $\gamma \in (0, 1]$ is the discount factor\cite{seid2021multi}. At each time step $t$, the agent observes a state $S_{t}\in \mathcal{S}$, then selects an action $A_{t}\in \mathcal{A}$, and receives a reward $R_{t}=R(S_{t},A_{t})$. The goal of the agent is to learn a policy that maximizes $\mathbb{E}\Big[\sum_{t=0}^{\infty}\gamma^t R_t\Big]$. 

In the context of wireless communication resource allocation, DRL has emerged as a promising solution to address the complexity and dynamics of wireless networks. The state space $\mathcal{S}$ typically encompasses various network parameters and channel conditions, including Channel State Information (CSI)\cite{zhang2022drl}, Queue State Information (QSI)\cite{hao2022delay}, user demands\cite{kasi2022d}, and buffer status\cite{swistak2024qos}. These state variables provide a comprehensive representation of the network environment, enabling the DRL agent to make informed decisions.
The action space $\mathcal{A}$ in wireless resource allocation generally consists of resource allocation decisions, such as power allocation levels, subcarrier assignments, bandwidth allocation, and scheduling priorities. These actions directly influence the network performance and user experience. For instance, in a multi-user MIMO system, the action space might include beamforming vectors and user selection decisions\cite{al2022self}, while in an OFDMA system, it could comprise subcarrier-user assignments and power allocation across different subcarriers\cite{he2023reinforcement}.
The reward function $R$ is typically designed to reflect the optimization objectives of the wireless network. Common reward metrics include achievable sum rate, energy efficiency, spectrum efficiency, and Quality of Service (QoS) satisfaction levels\cite{zhi2022deep}. To balance multiple objectives, the reward function can be formulated as a weighted sum, which uses weight factors to flexibly adjusting the relative importance of different objectives based on the desired trade-offs and priorities in the specific application scenario\cite{chai2024drl}. This formulation allows the DRL agent to learn resource allocation policies that optimize multiple network performance metrics simultaneously while adhering to various system constraints.
\vspace{-8pt}
\subsection{Generative Diffusion Models}
GDMs operate by iteratively generating and refining samples through diffusion and denoising. Its robust exploratory capabilities enhance training efficiency and stability. We detail it as follows.

\subsubsection{Forward Process} The forward diffusion process can be modeled as a Markov chain with $P$ steps. Denote $X^{0}$ as the original data at step-$0$. At each diffusion step-$p$, $p \in P$, Gaussian noise with a variance of $\beta_p$ is added to $X^{p-1}$, yielding $X^{p}$ with the distribution $q(X^p|X^{p-1})$. This process is expressed as: 
\begin{equation}\label{eq:q(Xp|X{P-1})}
q(X^p|X^{p-1}) = \mathcal{N}(X^p; \mu_p = \sqrt{1-\beta_p}X^{p-1}, \Sigma_p = \beta_p\mathbf{I}), 
\end{equation} 
where $q(X^p|X^{p-1})$ is a normal distribution with mean $\mu_p$ and variance $\Sigma_p$, and $\mathbf{I}$ is the identity matrix, representing equal standard deviation $\beta_p$ across all dimensions.

The posterior probability from the original data $X^0$ to the noisy data $X^P$ can be formulated as: \begin{equation} \label{eq:q(Xp|X0)}
q(X^P|X^0) = \prod_{p=1}^{P} q(X^p|X^{p-1}). 
\end{equation} 
Therefore, $X^p$ can be obtained through $p$ iterations of sampling. However, a large $p$ increases computational cost, to avoid this, by defining $\alpha_p = 1 - \beta_p$, $X^p$ can be derived in a single sampling step as: 
\begin{equation} \label{eq:p(X{p}|X0} 
{X}^p \sim q(X^p \mid X^0) = \mathcal{N}\left(X^p; \sqrt{\bar{\alpha}_p} X^0, (1 - \bar{\alpha}_p) \mathbf{I}\right), \end{equation} 
where $\bar{\alpha}_p = \prod_{m=1}^{p} \alpha_m$ is the cumulative product of $\alpha_m$. Note that $\beta_p$ is a hyperparameter, which can be fixed to a constant or chosen under a schedule over $P$ steps~\cite{song2020denoising}. Based on $\beta_p$, it is easy to precompute $\alpha_p$ and $\bar{\alpha}_p$ for all steps.

\subsubsection{Reverse Process for Denoised Data}
The denoised data $\hat{X}^{0}$ is inferred by progressively removing noise through a reverse process, starting with a sample $X^P$ drawn from a standard normal distribution, $\mathcal{N}(0, \mathbf{I})$. However, estimating the conditional probability distribution $q(X^{p-1} \mid X^p)$ is challenging. To address this, we construct a parameterized model $\phi$, defined as: \begin{equation}\label{eq:p(X{p-1}|Xp} 
\phi(X^{p-1} \mid X^p) = \mathcal{N}\left( X^{p-1}; \mu(X^p, p, S_t, A_t), \tilde{\beta}_p \mathbf{I} \right), 
\end{equation} 
where $\tilde{\beta}_p = \frac{1 - \bar{\alpha}_{p-1}}{1 - \bar{\alpha}_p} \beta_p$ denotes the deterministic variance, which can be derived from the known $\bar{\alpha}_p$ and $\beta_p$. Eq. (\ref{eq:p(X{p-1}|Xp}) allows us to obtain the trajectory from $X^P$ to $X^0$ as: 
\begin{equation} \label{eq:p{X0}}
\phi(X^0) = \phi(X^P)\prod_{p=1}^{P} \phi(X^{p-1} \mid X^p). 
\end{equation} 
By conditioning the model on diffusion step-$p$, it learns to predict the Gaussian parameters, i.e., the mean $\mu(X^p, p, S_t, A_t)$ and the covariance matrix $\tilde{\beta}_p \mathbf{I}$ for each diffusion step. Eq. (\ref{eq:p(X{p-1}|Xp}) can be modeled as a noise prediction model with the covariance matrix fixed as $\tilde{\beta}_p$, and
the estimated mean expressed as:
\begin{equation} \label{eq:mu}
\mu(X^p, p, S_t, A_t) = \frac{1}{\sqrt{\alpha_p}} \left( X^p - \frac{\beta_p}{\sqrt{1 - \bar{\alpha}_p}} \epsilon_{\Theta}(X^p, p, S_t, A_t) \right), \end{equation} 
where $\epsilon_{\Theta}(X^p, p, S_t, A_t)$ is a neural network with parameters $\Theta$ to generate denoising noise based on $S_t$ and $A_t$. 
We first sample ${X}^P \sim \mathcal{N}(0, \mathbf{I})$, then sample from the reverse diffusion chain parameterized by $\Theta$ as: 
\begin{equation} \label{X{p-1}|Xp}
X^{p-1} \mid X^p = \frac{X^p}{\sqrt{\alpha_p}} - \frac{\beta_p}{\sqrt{\alpha_p(1 - \bar{\alpha}_p)}} \epsilon_{\Theta}(X^p, p, S_t, A_t) + \sqrt{\beta_p} \epsilon, 
\end{equation} 
where $\epsilon \sim \mathcal{N}(0, \mathbf{I})$ and $p = 1, \ldots, P$. Furthermore, the loss function can be expressed as~\cite{du2024enhancing}:
\begin{equation} \label{eq: L}
\mathcal{L}_p = \mathbb{E}_{{X}_0, p, \epsilon} \left[ \Vert\epsilon - \epsilon_{\Theta} \left( \sqrt{\bar{\alpha}_p} {X}^0 + \sqrt{1 - \bar{\alpha}_p} \epsilon, p \right) \Vert^2 \right]. 
\end{equation} 
Eq. (\ref{eq: L}) shifts the model's task from predicting the mean of the distribution to predicting the noise $\epsilon$ at each diffusion step $p$.
\vspace{-6pt}
\section{System Model and Problem Formulation}\label{sec:system}
In this section, we present the system model of an FD mobile wireless communication system. In addition, we formulate the problem using MDP to facilitate the application of our proposed framework.
\vspace{-8pt}
\subsection{System Model}
We consider a network where an FD-based BS, equipped with two uniform linear arrays (ULAs), receives communication signals from $L$ single antenna
uplink users while simultaneously transmitting a downlink signal over the same time-frequency resources~\cite{he2023full}. The downlink signal, sent via a $N_t$-element ULA, serves to communicate with $K$ single-antenna downlink users. The uplink communication signals are captured by the receive ULA with $N_r$ elements at the BS.

In the system's downlink transmission, a narrowband signal is transmitted for multi-user communication using multi-antenna beamforming. The transmitted signal is formulated as
$ \mathbf{x} = \sum_{k=1}^{K} \mathbf{v}_k s_k$,
where $\mathbf{v}_k \in \mathbb{C}^{N_t \times 1}$ represents the beamforming vector for downlink user $k$ (with $k \in \{1, \cdots ,K\}$), and $s_k \in C$ is the data symbol for user $k$, with unit power, i.e., $\mathbb{E} \left\{ |s_k|^2 \right\} = 1$.
Furthermore, the total transmit power is subject to the constraint
$\sum_{k=1}^{K} \|\mathbf{v}_k\|^2 \leq P_{\text{max}}$, where $P_\text{max}$ represents the
maximum available power budget of the BS.

When the BS transmits $\mathbf{x}$, it simultaneously receives the uplink communication signals. Let
$u_l \in \mathbb{C}$ represent the uplink signal transmitted by user $l$ (with $l \in \{1, \cdots , L\}$),
which satisfies $\mathbb{E} \left\{ |u_l|^2 \right\} = p_l, \ \forall l$, where $0 \leq p_l \leq P_l$ denotes the average transmit power of user $l$, and $P_l$ is the maximum power budget of user $l$. The uplink channel between the $l$-$\text{th}$ user and the BS is represented by $h_l \in \mathbb{C}^{N_r \times 1}$, and the received multiuser uplink signal at the BS is expressed as $\sum_{l=1}^{L} \mathbf{h}_l d_l$. The uplink transmission design is carried out by adjusting the transmit power $\{ p_l \}_{l=1}^{L}$ of uplink users.

Assuming that both the transmit and receive ULAs at the BS have half-wavelength antenna spacing, we define the transmit array steering vector in the direction of $\theta$ as $\mathbf{a}_t(\theta) \triangleq \frac{1}{\sqrt{N_t}} \left[ 1, e^{j\pi \sin(\theta)}, \dots, e^{j\pi (N_t-1) \sin(\theta)} \right]^T$. Similarly, the receive steering vector is defined as $\mathbf{a}_r(\theta) \triangleq \frac{1}{\sqrt{N_r}} \left[ 1, e^{j\pi \sin(\theta)}, \dots, e^{j\pi (N_r-1) \sin(\theta)} \right]^T$. 
Given the uplink communication signal, the received signal at the FD-based BS is expressed as:
\begin{equation}
    \mathbf{y}^{\text{BS}} = \sum_{l=1}^{L} \mathbf{h}_l d_l + \mathbf{z} + \mathbf{n},
\end{equation}
where 
$\mathbf{n} \in \mathbb{C}^{N_r \times 1}$
denotes additive white Gaussian noise (AWGN) with covariance $\sigma_r^2 \mathbf{I}_{N_r}$, and $\mathbf{z} \in \mathbb{C}^{N_r \times 1}$ denotes the signal-dependent interference, which can be divided into two parts. The first part represents the self-interference (SI) caused by the FD operation, which can be mitigated using SI cancellation (SIC) techniques within the FD system\cite{chauhan2024full}. The second part corresponds to the clutter reflected
from the surrounding environment. Assuming that there exist $F$ signal-dependent uncorrelated interferers located at angles $\{\theta_f\}_{f=1}^F$, $\forall f \in \{1, \dots, F\}$. These $F$ interferers also reflect the sensing signal to the BS, yielding the undesired interference $\sum_{f=1}^F \beta_f \mathbf{A(\theta_f)} \mathbf{x}$ with $\beta_f \in \mathbb{C}$ being the complex amplitude of the $f$-th interferer and $\mathbf{A(\theta_f)} = a_r(\theta_f) a_t^H(\theta_f)$, $\forall f$~\cite{he2023full}.

Denote the channel between downlink user $k$ and the BS as $g_k \in \mathbb{C}^{N_t \times 1}$. The received signal at downlink user $k$ is then expressed as
\begin{equation}
    y_k^{\text{User}} = {\mathbf{g}_k^H \mathbf{v}_k s_k} + n_k, \quad \forall k,
\end{equation}
where $n_k$ represents the AWGN with variance $\sigma_k^2$. Given that the interference from other users is effectively suppressed due to advanced interference mitigation techniques, the signal-to-noise ratio (SNR) for downlink user $k$ is given by 
\begin{equation}
    \gamma_k^{\text{com, DL}} = \frac{|\mathbf{g}_k^H \mathbf{v}_k|^2}{\sigma_k^2}, \quad \forall k.
\end{equation}
To recover the signal from uplink at the BS, a set of receive beamformers $\left\{ \mathbf{w}_l \right\}_{l=1}^{L} \in \mathbb{C}^{N_r \times 1}$ are applied to the received signal $\mathbf{y}^{\text{BS}}$. The corresponding signal-to-interference-plus-noise ratio (SINR) for uplink user $l$ is expressed as~\cite{he2023full}:
\begin{equation}
    \gamma_l^{\text{com, UL}} = \frac{p_l \mathbf{w}_l^H \mathbf{h}_l \mathbf{h}_l^H \mathbf{w}_l}
{\mathbf{w}_l^H \left( \sum_{l' = 1, l' \neq l}^{L} p_{l'} \mathbf{h}_{l'} \mathbf{h}_{l'}^H + \mathbf{GQG}^H + \sigma_r^2 \mathbf{I}_{N_r} \right) \mathbf{w}_l}, \quad \forall l,
\end{equation}
where $\mathbf{G} = \sum_{f=1}^F\beta_f \mathbf{A}(\theta_f)$ denotes the interference channel from the environment, and $\mathbf{Q} = \mathbb{E} \left\{ \mathbf{x} \mathbf{x}^H \right\} = \sum_{k=1}^{K} \mathbf{v}_k \mathbf{v}_k^H$ denotes the covariance matrix of the downlink signal.

The objective is to maximize the total sum rate of all the uplink and downlink users, subject to the constraint of limited transmit power budgets. The optimization problem can be formulated as follows:
\begin{subequations}\label{eq: max-sum-rate} 
\begin{align}
    (\textbf{P0}) &\max_{\mathcal{O}} \quad C = \sum_{l=1}^{L} \log_2 \left( 1 + \gamma_l^{\text{com, UL}} \right) + \sum_{k=1}^{K} \log_2 \left( 1 + \gamma_k^{\text{com, DL}}\right) \\
    \quad & \text{s.t.} \quad \sum_{k=1}^{K} \|\mathbf{v}_k\|^2 \leq P_{\text{max}}, \\
    & \quad 0 \leq p_l \leq P_l, \quad \forall l,
\end{align}
\end{subequations}
where $\mathcal{O}=\{\left\{\mathbf{w}_l \right\}_{l=1}^{L}, \left\{ \mathbf{v}_k \right\}_{k=1}^{K}, \left\{p_l \right\}_{l=1}^{L}\}$ is the set of
optimization variables. 
\vspace{-8pt}
\subsection{Problem Reformulation}
Conventional optimization techniques for addressing the resource allocation problem often necessitate accurate CSI. However, obtaining precise CSI incurs high computational costs, resulting in substantial resource utilization and latency. These drawbacks are especially critical in real-time communication networks characterized by highly dynamic and rapidly fluctuating environments. To overcome these limitations, we reformulate the resource allocation problem as an MDP.
\subsubsection{State Space, Action Space, and Reward Function}
The state space of the BS at each time step is $\mathcal{S} = \{\{\left(x_k,y_k\right)\}_{k=1}^{K}, \{\left(x_l,y_l\right)\}_{l=1}^{L}, \{\mathbf{g}_k\}_{k=1}^{K}, \{\mathbf{h}_l\}_{l=1}^{L}\}$, where $(x_k, y_k)$ is the location of downlink user $k$ at current time step, $(x_l, y_l)$ is the location of uplink user $l$ at current time step, $\mathbf{g}_k$ and $\mathbf{h}_l$ denote the downlink channel gain for user $k$ and uplink channel gain for user $l$, respectively. 
The action space of the BS at each time step corresponds to the optimization variables in problem $\textbf{P0}$, expressed as $\mathcal{A} = \{\left\{ \mathbf{w}_l \right\}_{l=1}^{L}, \left\{ \mathbf{v}_k \right\}_{k=1}^{K}, \left\{ p_l \geq 0 \right\}_{l=1}^{L}\}
$. 
The most straightforward expression for the reward function would be to use the optimization objective. In this network, the reward function can be directly designed as the sum rate $R=C$.



\subsubsection{Optimization Formulation}
Let $\mathcal{\pi}$ denote a stochastic policy (i.e., $\mathcal{\pi}: \mathcal{S} \times \mathcal{A} \rightarrow [0, 1]$) which is the probability that action $A_t$ is taken at time step $t$ given the state $S_t$, i.e., $\mathcal{\pi} = \text{Pr}\{A_t|S_t\}$. Given the discount factor $\gamma$, let $J(\mathcal{\pi})$ denote the expected discounted reward of the network by following policy $\mathcal{\pi}$:
\begin{equation}
\label{eq:discounted-reward}
J(\mathcal{\pi}) = \mathbb{E}_{A_t \sim \mathcal{\pi}, S_t \sim \mathcal{P}} \Big[\sum_{t=0}^{\infty}\gamma^t R_t(S_t, A_t)\Big].
\end{equation}
Our goal is to find the optimal policy $\mathcal{\pi}^*$ for the network that maximizes $J(\mathcal{\pi}$), i.e., 
\begin{equation}
\label{eq:max-return}
\begin{aligned}
\max_{\mathcal{\pi}} \quad & J(\mathcal{\pi})\\
\textrm{s.t.} \quad & a_t \sim \mathcal{\pi}(A_t|S_t), S_{t+1} \sim \mathcal{P}(S_{t+1}|S_t, A_t),
\end{aligned}
\end{equation}
where $\mathcal{P}(S_{t+1}|S_t, A_t)$ is the transition process from $S_t$ to $S_{t+1}$.
\vspace{-6pt}
\section{Challenges and Proposed Method}
\begin{figure*}[t]
  \centering
  \includegraphics[width=.9\textwidth]{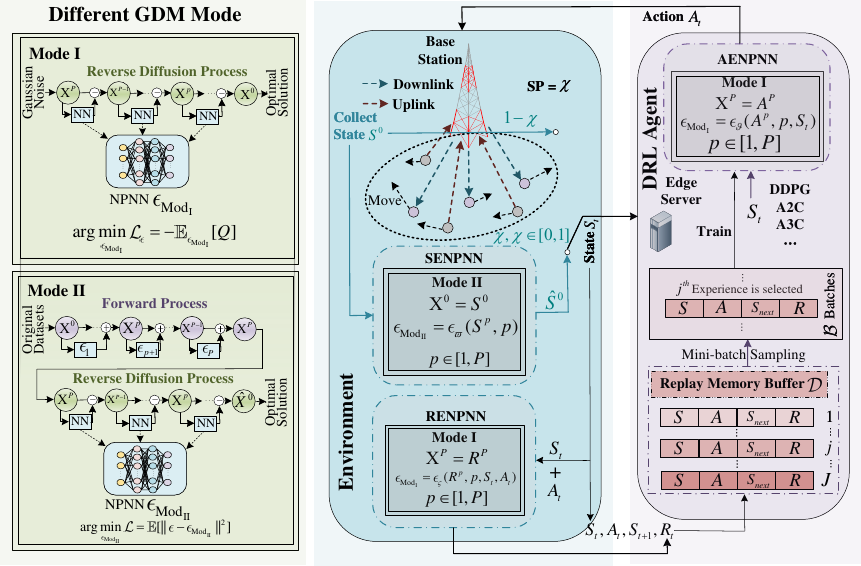}\\
  \caption{\textcolor{black}{Illustration of D2RL Framework. }}
  \label{fig:Archi}
\end{figure*}

In this section, we discuss the challenges encountered when applying DRL to mobile wireless communication systems. We then present the gradient analysis of the optimization problem. Finally, we introduce our proposed D2RL framework to address the identified challenges and effectively solve the problem aforementioned.
\label{sec:algo}
\vspace{-8pt}
\subsection{Challenges of Applying DRL in Wireless Communications}
The application of DRL in wireless communications presents several fundamental challenges across its key components - state space, action space, and reward space. These challenges significantly impact the training efficiency and performance of DRL systems in practical wireless network deployments.
\subsubsection{Challenges for State Data Collection}
In wireless communication networks, the state space encompasses critical information such as CSI, QSI, and user demands, presenting significant challenges in both acquisition and management. The collection of accurate state information, particularly CSI, demands substantial computational resources and introduces considerable network overhead, especially in dynamic environments. This leads to limited training samples, impeding DRL agents' ability to comprehend the environment and make optimal decisions. Moreover, the high-dimensional, time-varying nature of wireless network states further complicates learning and generalization. This combination of limited training data and complex state representations particularly affects the agent's ability to generalize to unseen network conditions, potentially compromising the system's performance in real-world deployments.
Notably, GDMs offer a promising solution by generating realistic synthetic state samples, augmenting limited training data while preserving wireless channel statistics.

\subsubsection{Challenges for Action Space Exploration}
The action space in wireless communication resource allocation presents significant challenges that compound the complexity of decision-making processes. The continuous nature of wireless resource allocation variables, such as power levels and bandwidth, complicates action space exploration and optimization for DRL agents. Stringent real-time requirements necessitate swift yet optimal actions while satisfying multiple constraints, including power budgets and QoS, creating a complex multifaceted optimization problem. In this context, GDMs can provide an innovative solution by learning the underlying distribution of optimal actions, enabling more efficient exploration of the continuous action space while maintaining feasibility within the network constraints.
\subsubsection{Challenges for Reward Function Design}
Designing effective reward functions for wireless communications is challenging due to competing objectives like throughput, energy efficiency, and latency. Encapsulating diverse requirements into a single function is complex. While expert knowledge can provide valuable insights into reward function design, an overreliance on predetermined reward structures may inadvertently constrain the exploration space and limit the discovery of optimal solutions. This limitation is particularly evident in complex wireless scenarios where expert-designed rewards might fail to capture subtle trade-offs, potentially leading to convergence toward local optima rather than optimal solutions. Furthermore, the challenge is exacerbated by the temporal aspects of wireless networks, where the consequences of specific actions may manifest with considerable delay, complicating the design of reward functions.
GDMs offer a practical approach by learning implicit relationships between actions and consequences, enabling nuanced and adaptive reward structures that capture wireless networks
\vspace{-8pt}
\subsection{The Gradient Analysis of the Optimization Problem}
Following the MDP formulation presented in Section \ref{sec:system} B, we investigate the gradient properties of the objective function in Eq. (\ref{eq:discounted-reward}) to analyze the training efficiency of DRL.

To derive the gradient of $J(\pi)$, we employ the score function estimator (SFE) (also used as the likelihood ratio trick), which provides a fundamental relationship for computing gradients of expectations:
\begin{equation}\label{eq:principle}
	 \nabla \mathbb{E}_{x \sim p} [f(x)] = \mathbb{E}_{x \sim p} \left[ f(x) \nabla \log p(x) \right], 
\end{equation}
where $p(x)$ denotes the probability density function and $f(x)$ represents the function being evaluated under the expectation.

In the context of policy optimization, we consider trajectories $\tau = (S_0, A_0, R_0, S_1, A_1, R_1, \dots)$ generated by policy $\pi$ under the environment dynamics. The probability over trajectories, assuming the Markov property, follows as:
\begin{equation} \label{eq:trajectory}
	p(\tau) = p(S_0) \prod_{t=0}^\infty \pi(A_t | S_t) \mathcal{P}(S_{t+1} | S_t, A_t), 
\end{equation}
where $p(S_0)$ denotes the initial state distribution, $\pi(A_t | S_t)$ represents the policy's action probability distribution, and $\mathcal{P}(S_{t+1} | S_t, A_t)$ describes the environment's transition dynamics. Note that we consider a finite horizon $T$ for practical implementation, though the theoretical framework extends to the infinite horizon case under appropriate regularity conditions.

The objective function can be expressed as an expectation over trajectories:
\begin{equation} 
	J(\pi) = \int_\tau p(\tau) \sum_{t=0}^\infty \gamma^t R_t(S_t, A_t). 
\end{equation}
Under suitable regularity conditions that ensure the interchange of differentiation and integration (Leibniz integral rule), the gradient of $J(\pi)$ can be derived as follows:

\begin{equation} 
	\nabla J(\pi) = \int_\tau \nabla p(\tau) \sum_{t=0}^T \gamma^t R_t(S_t, A_t).
\end{equation}
Applying the SFE and noting that the environment dynamics $\mathcal{P}$ are independent of policy parameters, we obtain:

\begin{equation} \label{eq: grad_deriv}
\nabla J(\pi) = \mathbb{E}_{\pi} \left[ \sum_{t=0}^T \nabla \log \pi(A_t | S_t) G_t \right],
\end{equation}
where $G_t = \sum_{k=t}^T \gamma^{k-t} R_k$ represents the discounted cumulative return from time step $t$.

The derived gradient formulation elucidates several fun-
damental factors that substantially influence the training ef-
ficiency of DRL. The analysis of Eq. (\ref{eq: grad_deriv}) reveals that the policy gradient magnitude is directly proportional to the log-probability gradients of the policy's action selection and the corresponding cumulative returns. Fluctuations in the state-action space topology or reward distribution can substantially impact gradient updates, affecting DRL training efficiency and stability. This motivates investigating the action space, state space, and reward space to determine the learning process's efficiency and stability.
\vspace{-8pt}
\subsection{Proposed Diffusion-based DRL (D2RL) Framework}
The proposed D2RL framework systematically explores the action, state, and reward spaces to guide the agent in making more informed and efficient decisions, ultimately optimizing its performance.

The overall architecture of D2RL is shown in Fig.~\ref{fig:Archi}.
\subsubsection{Different Modes of GDMs}Based on the accessibility of the original dataset, the two different GDMs modes can be proposed to the explore these three spaces in DRL. 
\begin{itemize}
    \item 
    Mode I: 
Mode I is specifically designed for scenarios where an original dataset is unavailable, relying solely on the reverse process of GDM. As illustrated in Fig.~\ref{fig:Archi}, Mode I begins by generating random Gaussian noise as the input for step-$P$. Subsequently, the noise predictor Neural Network (NPNN) is iteratively utilized to estimate the noise at the current step. This estimated noise is then subtracted from the data at step-$p$ to derive the data for step-$(p-1)$, where $p \in [1,P]$. This process continues iteratively until the denoised data at step-$0$, denoted as $X^0$, is obtained. The objective of the NPNN for Mode I is mathematically formulated as:
\begin{equation}
\arg\min_{\epsilon_{\text{Mod}_\text{I}}} \mathcal{L}_\epsilon = -\mathbb{E}_{\epsilon_{\text{Mod}_\text{I}}} \left[ Q \right],
\end{equation}
which aims to optimize the Q-value.

Specifically, Mode I is used to explore the action and reward spaces, facilitating the agent's learning process in the absence of an original dataset. This approach enables the agent to gradually refine its decision-making process through reverse exploration, ultimately guiding it towards more efficient actions and better reward expressions.
\item 
    Mode II: 
Mode II is applicable when an original dataset is available. It involves a multi-step forward proceduress applied to the original data, followed by a reverse process using the noise predictor to generate new data. The generated data can then be used to augment the original dataset. For instance, at step-$p$ of the forward process, the sum of $X^p$ and randomly generated noise $\epsilon_{p+1}$ serves as the noisy data $X^{p+1}$ for step-$(p+1)$, $p \in [0,P-1]$. Upon completion of the noise addition, the noisy data from step-$P$ serves as the initial input for the reverse process. This process follows the same steps as in Mode I to generate the data, denoted as $\hat{X}^{0}$. The optimization objective of the NPNN in Mode II can be mathematically represented as:
\begin{equation}
\arg\min_{\epsilon_{\text{Mod}_\text{II}}} \mathcal{L} = \mathbb{E} \left[ \left\| \epsilon - \epsilon_{{\text{Mod}_\text{II}}} \right\|^2 \right],
\end{equation} 
which aims to minimize the difference between the predicted noise during the reverse process and the noise added during the forward process. 

Mode II is employed to explore the state space by leveraging both the original and generated data, enhancing the agent's decision-making through comprehensive state exploration.
\end{itemize}
 
We now provide a comprehensive overview of the D2RL framework. As depicted in Fig.~\ref{fig:Archi}, the framework consists of two primary components: the environment and the DRL agent. On the left side, the environment is coupled with the proposed reward exploration network\footnote{\textcolor{black}{In practical applications, this can be executed in the environment (i.e., the computational unit) before being sent to the agent.}} and the state exploration network. On the right side, the DRL agent interacts with the environment. To be noticed, the DRL can be any algorithm (e.g., DDPG, A2C, A3C). For illustrative purposes, we use DDPG as the baseline algorithm. The DDPG framework consists of several key components that collaborate to optimize the agent's strategy. 

\subsubsection{State Exploration Network}
To accelerate the learning process and uncover information that current state data may not capture, we propose utilizing the existing state dataset to generate new data using Mode II. Our approach involves training a \textcolor{black}{State Exploration NPNN (SENPNN)}, denoted as $\epsilon_{\varpi}$, using the original state information $S^0$ collected from the environment at each time step. The SENPNN generates an explored state $\hat{S}^{0}$ through Mode II, which is then integrated into the DRL training process.
During the training phase, the agent selects either the newly generated state $\hat{S}^{0}$ with a substitute probability (SP) of $\chi$ or the original state $S^0$ with a probability of $1-\chi$ as the current state information $S_t$ received from the environment. Based on the selected state $S_t$, the agent generates a corresponding action $A_t$ and provides feedback to the environment, which in turn yields a reward $R_t$ based on the action $A_t$.

It is crucial to acknowledge that during the early stages of SENPNN training, the newly generated state $\hat{S}^{0}$ may not closely resemble the original state $S^0$. Consequently, a high probability of selecting $\hat{S}^{0}$ during this phase can severely impair the training efficiency of the DRL model. To mitigate this issue, we propose a training mechanism that gradually increases the value of $\chi$. Initially, $\chi$ is set to 0, and as the training epochs progress, $\chi$ grows at an update rate of $\eta$ when the loss function value of the SENPNN falls below a predefined threshold $\mathcal{T}$. Furthermore, we define a maximum replacement probability $\mathcal{M}$ for $\hat{S}^{0}$, which satisfies the condition $\chi=\min(\chi+\eta,\mathcal{M})$.


The proposed methodology enables the DRL model to leverage the benefits of GDM-based state exploration while maintaining the stability and effectiveness of the training process. By incorporating the explored states $\hat{S}^{0}$ generated by the SENPNN, the agent can uncover valuable information that may not be captured by the current state data, leading to enhanced performance and faster convergence of the DRL model. The complete algorithm for GDM-based state exploration is presented in \textbf{Algorithm \ref{alg:SENPNN_Exploration}}.
   
\subsubsection{Action Exploration Network} 
When the agent receives the state $S_t$ from the environment, as shown in Fig.~\ref{fig:Archi}, it generates an action $A_t$ by employing the GDM exploration method of Mode I. The process involves training an Action Exploration NPNN (AENPNN), denoted as $\epsilon_{\vartheta}$, which takes a combination of the current state $S_t$ and a randomly generated Gaussian noise $A^P$ as the input.
The AENPNN operates by first concatenating the current state $S_t$ with the Gaussian noise $A^P$. This combined input then undergoes $P$ steps of the GDM reverse process, which is a critical component of the exploration method. The GDM reverse process iteratively refines the input, allowing the AENPNN to generate an action $A_t$ that is based on the current state $S_t$ but also incorporates exploratory behavior induced by the Gaussian noise $A^P$.

\subsubsection{Reward Exploration Network}
When the environment receives the agent's action $A_t$ based on state $S_t$, it evaluates the action and assigns a score, thereby providing feedback to the agent for generating better actions in the future. As depicted in Fig.~\ref{fig:Archi}, we employ the GDM exploration method of Mode I to generate the reward $R_t$, which serves as a critical component of the agent's learning process. The reward generation process involves training a Reward Exploration NPNN (RENPNN), denoted as $\epsilon_{\varsigma}$. The RENPNN takes a combination of the current state $S_t$, the action $A_t$, and a randomly generated Gaussian noise $R^P$ as input. By incorporating the state-action pair $(S_t, A_t)$ along with the Gaussian noise $R^P$, the RENPNN can generate rewards that consider both the current context and exploratory variations. The input to the RENPNN undergoes $P$ steps of the GDM reverse process, which is a key aspect of the exploration method. The GDM reverse process iteratively refines the input, allowing the RENPNN to generate a reward $R_t$ that is conditioned on the current state $S_t$ and action $A_t$ while incorporating exploratory behavior induced by the Gaussian noise $R^P$. The generated reward $R_t$, along with the new state $S_{t+1}$ produced by the environment based on the agent's action, is then fed back to the agent. This feedback loop enables the agent to learn and adapt its behavior over time, seeking to maximize the cumulative rewards obtained during the interaction with the environment.

\begin{algorithm}[t]
   \caption{\textcolor{black}{GDM-based State Exploration Algorithm}}
   \label{alg:SENPNN_Exploration}
\begin{algorithmic}
\STATE {\bfseries Input:} original state $\boldsymbol{S}^0$, GDM exploration strategy Mode II, SENPNN parameters $\varpi$, update rate $\eta$, maximum replacement probability $\mathcal{M}$, diffusion step $P$, initial SP $\chi=0$.
\STATE Set the loss threshold for SENPNN training $\mathcal{T}$.
\STATE Generate $\boldsymbol\hat{S}^{0}$ using $\boldsymbol{S}^0$ and Mode II.
    \IF{SENPNN loss $\mathcal{L}(\boldsymbol{S}^0, \boldsymbol\hat{S}^{0}) < \mathcal{T}$}
        \STATE Increase exploration probability: $\chi \leftarrow \min(\chi + \eta, \mathcal{M})$.
    \ENDIF
\STATE Select current state $\boldsymbol{S}_t$ based on $\chi$:
    \IF{random number $r \leq \chi$}
        \STATE Use the explored state: $\boldsymbol{S}_t = \boldsymbol\hat{S}^{0}$.
    \ELSE
        \STATE Use the original state: $\boldsymbol{S}_t = \boldsymbol{S}^0$.
    \ENDIF
\STATE Generate corresponding action $A_t$ based on $S_t$.
\STATE Provide feedback to the environment to get reward $R_t$.
\end{algorithmic}
\end{algorithm}
\setlength{\textfloatsep}{3pt} 
When the agent receives $S_{t+1}$ and $R_t$, it records $(S_t,A_t,S_{t+1},R_t)$ into the replay memory buffer. The above steps are iterated until the agent learns the policy.
At each time step, the agent takes a mini-batch $\mathcal{B}=
(S_j,A_j,S_{j+1},R_j)$ of data at a time from the replay memory buffer to update the policy\cite{yu2024attention}. In order to mitigate the problem of overestimation bias of single critic, D2RL framework uses double critic network, which consists of two separate critic networks, $Q_{1,\xi_1}$ and $Q_{2,\xi_2}$, each responsible for independently estimating the Q-value, which represents the expected cumulative reward given the current state and action:
\begin{equation}\label{eq: Q}
    Q(S_j, A_j) = \mathbb{E}\left[R + \gamma \cdot Q(S_{j+1}, A_{j+1})\right],
\end{equation}
where $\gamma$ is the discount factor used to measure the importance of future rewards.
During training, both critic network parameters $\xi=\{\xi_1,\xi_2\}$ are updated, but the minimum of the two Q-values is chosen to update the actor network, which reduces overestimation by ensuring that the agent relies on the more conservative Q-value estimate.

In addition, D2RL framework uses target networks to enhance training stability by freezing their parameters during gradient descent and updating them slowly via a soft update mechanism, denoted as
\begin{equation}\label{eq:update A, Q}
    \begin{aligned}
        &\hat{\vartheta}_{e+1} \leftarrow \tau \vartheta_e + (1 - \tau)\hat{\vartheta}_e,\\
        &\hat{\xi}_{e+1} \leftarrow \tau \xi_e + (1 - \tau) \hat{\xi}_e,
    \end{aligned}
\end{equation}
where $\hat{\vartheta}_e$ and $\hat{\xi}_e$ are the parameters of the target actor network and target critic network during the $e$-th epoch, respectively, and $\tau \in (0, 1]$ controls the frequency at which the target network is updated\cite{hu2023intelligent}.

The pseudocode of the algorithm proposed for the D2RL framework is shown in \textbf{Algorithm 2}.
\begin{algorithm}[!t]
\label{algo2:D2RL}
 \caption{\textcolor{black}{Diffusion-based DRL (D2RL)}}
\begin{algorithmic}
\STATE {\bfseries Initiate:} Initialize critic-network parameters $\xi_1$ and $\xi_2$, target-critic-network parameters $\hat{\xi} \leftarrow \min(\xi_1,\xi_2)$, SENPNN parameters $\varpi$, AENPNN parameters $\vartheta$, target-AENPNN parameters $\hat{\vartheta} \leftarrow \vartheta$, RENPNN parameters $\varsigma$, and experience replay buffer $\mathcal{D}$.
\STATE {\bfseries Input:} GDM exploration strategy Mode I, II.
\FOR {the training epoch $e = 1$ to $E$}
\FOR{the collected experiences $t = 1$ to $T$}
\STATE{Observe the state ${S^0}$ from the environment}.
\STATE{Obtain $S_t$ using \textbf{Algorithm \ref{alg:SENPNN_Exploration}}, action ${A_t}$ based on Mode I and $S_t$, reward ${R_t}$ based on Mode II, $S_t$ and ${A_t}$.}
\STATE{Obtain the next state ${S_{t+1}}$.}
\STATE{Record experience tuple $(S_t, A_t, S_{t+1}, R_t)$ in replay memory buffer $\mathcal{D}$.}
\ENDFOR
\STATE{Extract a batch of experiences $\mathcal{B} = (S_j, A_j, S_{j+1}, R_j)$ from the experience replay buffer $\mathcal{D}$ for optimization;}
\STATE{Update the SENPNN parameters $\varpi$, AENPNN parameters $\vartheta$, double critic network parameters $\xi_1$ and $\xi_2$, RENPNN parameters $\varsigma$.}
\STATE{Perform a soft update of the target network parameters $\hat{\vartheta}, \hat{\xi}$ using Eq. (\ref{eq:update A, Q}).}
\ENDFOR
\STATE{Return the optimal policy.}
\end{algorithmic}
\end{algorithm}
\vspace{-6pt}
\section{Evaluation}
\label{sec:evaluation}
In this section, we evaluate the performance of the proposed D2RL framework in addressing the total data rate maximization problem.

\begin{table}[ht]
\centering
\caption{Environment setup}
\label{table1}
\begin{tabular}{l l}
\hline
\textbf{Parameters} & \textbf{Value} \\ \hline
Cell radius $M$ & 300 m \\ 
Number of downlink users $K$ & 6 \\ 
Number of uplink users $L$ & 4 \\ 
Height of BS  & 100 m \\ 
Pathloss exponent $\alpha$ & 3.6 \\ 
Noise power $\sigma_k^2, \sigma_r^2$& -97 dBm/Hz \\ 
Number of transmit antenna $N_t$ & 6 \\
Number of receive antenna $N_r$ & 6 \\
Channel gain $\rho_0$ & $4.16 \times 10^{-6}$ \\  
Maximum BS transmit power $P_\text{max}$ & 3 W \\ 
\makecell[l]{Maximum transmit power of\\ uplink users $\{P_l\}_{l=1}^L$} & 1 W \\
\hline
\end{tabular}
\end{table}

\begin{table}[ht]
\centering
\caption{Simulation parameters for the proposed D2RL}
\label{table2}
\begin{tabular}{l  l  l  l }
\hline
\textbf{Parameters} & \textbf{Value}&\textbf{Parameters} & \textbf{Value} \\ \hline
\makecell[l]{HL number\\ of AENPNN \\ $\&$ SENPNN}  & 4 & \makecell[l]{HL number\\of RENPNN} &5\\
\makecell[l]{Number of HL\\ neurons} &256 & Weight decay & $7 \times 10^{-5}$ \\
\makecell[l]{Learning rate\\ of AENPNN \\ $\&$ RENPNN}  & $5 \times 10^{-5}$  &\makecell[l]{Learning rate\\ of SENPNN}  & $1 \times 10^{-4}$ \\
$\mathcal{T}$ &$5 \times 10^{-4}$  &$\epsilon$-greedy $\epsilon$ & 0.1 \\ 
Soft update $\tau$ & $5 \times 10^{-3}$ 
&$\gamma$ & 1  \\ 
Diffusion steps  & 6  
&Batch size $\mathcal{B}$ & 256 \\ 
Buffer size & $10^5$ 
&Optimizer & ADAM \\ \hline
\end{tabular}
\begin{tablenotes}
\small
\item HL: Hidden Layer
\end{tablenotes}
\vspace{-.1cm}
\end{table}

\vspace{-8pt}
\subsection{Experimental Settings}

\subsubsection{Simulation Setting}
It is assumed that
$F=2$ interferers are located at $\theta_1=-50^{\circ}$ and $\theta_2=20^{\circ}$,
respectively. The channel gain of the two interferers are set to $\frac{|\beta_1|^2}{\sigma_r^2} = \frac{|\beta_2|^2}{\sigma_r^2} = 20 \ \text{dB}$~\cite{he2023full}. The channel model for the users is assumed to follow a line-of-sight (LoS) path. Specifically, for uplink users, the channel gain is modeled as $h_l=\rho_0d_l^{-\alpha}\mathbf{a}_r(\theta_l^{\text{UL}}), \forall l$, where $\rho_0$ is the channel gain at the reference distance $d_0=1~\text{m}$, $d_l$ is the distance between uplink user $l$ and the BS, $\alpha$ is the path-loss exponent, and $\theta_l^{\text{UL}}$ is the angle of user $l$ relative to the BS\cite{miao2024utility}. Similarly, for downlink users, the channel gain is represented as $g_k=\rho_0d_k^{-\alpha}\mathbf{a}_t(\theta_k^{\text{DL}}), \forall k$. A summary of the key simulation parameters is provided in Table \ref{table1}. The supplementary for the hyperparameter settings are in Table \ref{table2}. 

\subsubsection{Baseline Setting}
We adopt the model architecture proposed by \cite{du2024enhancing}, which explores the action space, and apply it to our FD mobile wireless communication network. To evaluate the effectiveness of the proposed approach, we conduct a comparative analysis of two distinct scenarios:
\begin{itemize}
    \item \textbf{A. w/o GDM}: The action is generated based on a traditional NN.
    \item \textbf{A. w/ GDM}: The action space is explored using GDMs.
\end{itemize}

\subsubsection{Reward Exploration Setting}
While preserving the AENPNN, we further investigate the effect of introducing the RENPNN on training performance. To provide a comprehensive evaluation of how different reward function designs influence training outcomes, we compare the training performance across $5$ reward function designs:
\begin{itemize}
    \item \textbf{Raw R.}: The reward directly uses the system objective $R=C$ \cite{nayak2024drl}.
    \item \textbf{Designed R.}: The reward function is designed under the guidance of expert knowledge (that is, theoretical upper bound) $R^{'}$ as $R=C-R^{'}$ \cite{du2024enhancing}.
    \item \textbf{Designed R. + MLP}: The designed reward function is adjusted and explored using a Multilayer Perception (MLP).
    \item \textbf{GDM R.}: The designed reward function is directly explored using GDMs.
    \item \textbf{Designed + GDM R.}: Based on the Designed R., the designed reward function is further enhanced and explored using GDMs.
\end{itemize}

\subsubsection{State Exploration Setting}
Building upon the foundation of action exploration and reward exploration, we extend our research to examine the impact on training performance when a portion of the original state data is replaced using GDMs. To systematically evaluate the influence of state exploration on training efficiency, we introduce two aforementioned hyperparameters: $\mathcal{M}$ and $\eta$. By varying these hyperparameters, we conduct a comparative analysis of system performance with and without state exploration under $5$ reward design scenarios:
\begin{itemize}
    \item \textbf{Raw R. + S. Exp.}: State exploration is introduced on top of the raw reward $R=C$.
    \item \textbf{Designed R. + S. Exp.}: State exploration is introduced on top of the designed reward function.
    \item \textbf{Designed R. + MLP + S. Exp.}: State exploration is introduced on top of the Designed R. adjusted by MLP.
    \item \textbf{GDM R. + S. Exp.}: State exploration is introduced on top of the GDM reward function.
    \item \textbf{Designed + GDM R. + S. Exp.}: State exploration is introduced on top of the Designed R. enhanced by GDMs.
\end{itemize}

\subsubsection{Evaluation Metric}
To assess the performance and effectiveness of the proposed approach, we employ a comprehensive set of evaluation metrics. 
\begin{itemize}
    \item \textbf{Sum Rate $\&$ Moving Averaging (MA) Reward\cite{zhou2024federated}}: The sum rate serves as the optimization objective of the problem, directly reflects the efficiency of resource allocation. Additionally, the MA reward provides a measure of the quality of resource allocation decisions over time. These two metrics offer a clear indication of the training efficiency achieved under different space exploration cases.
    \item \textbf{Gradient Weight $\&$ Gradient Bias}: To better understand the learning dynamics and performance variations across different space exploration scenarios, we analyze the gradient weight and bias, which are fundamental metrics that quantify the learning process's progression. Gradient weight is the primary driver of learning, optimizing feature extraction and input-output mappings while addressing challenges like vanishing or exploding gradients. Gradient bias, although smaller in magnitude, modulates neuronal activation thresholds and facilitates adaptation to diverse data distributions. When properly calibrated, gradient bias can accelerate initial training phases by enabling rapid threshold adjustments and helping networks overcome suboptimal initialization states. 
    \item \textbf{Total GPU Time $\&$ GPU Time Per Epoch}: Furthermore, we recognize the importance of computational efficiency in the training process. To this end, we evaluate the total GPU time and the GPU time per epoch. These metrics allow us to quantify the trade-off between the increased complexity introduced by the exploration techniques and the potential benefits of faster convergence. By considering both the performance improvements and the computational costs, we aim to provide a holistic assessment of the proposed approach and its practical viability.
\end{itemize}
\vspace{-8pt}
\subsection{Performance Evaluation}
\begin{figure}[t]
    \centering
    \includegraphics[width=.58\linewidth]{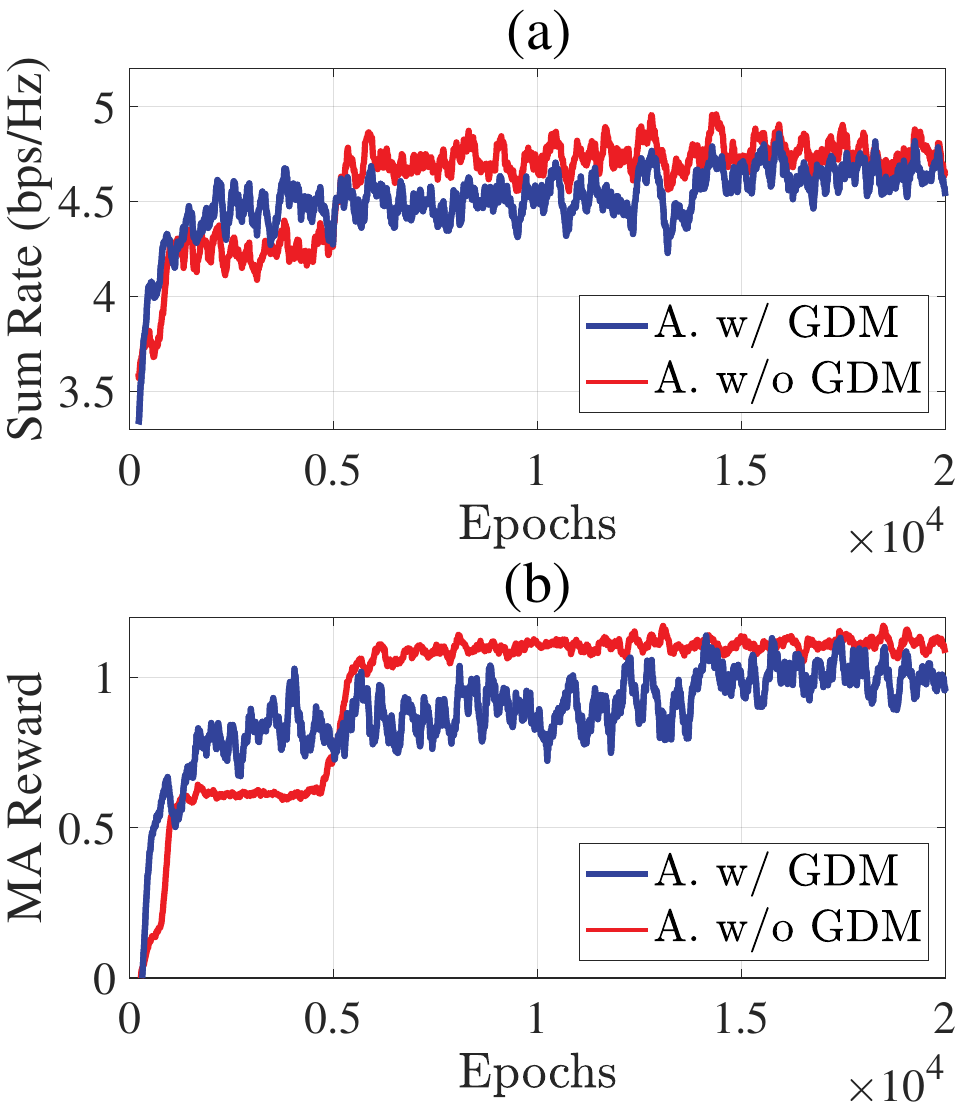}
    \caption{Comparison of (a) the sum rate achieved by GDM-based action exploration and without action exploration, and (b) MA reward obtained by GDM-based action exploration and without action exploration.
   }  
    \label{fig:sum_rate_reward_action_com}
\end{figure}
\vspace{-5pt}

\begin{figure}[!t]
    \centering
    \includegraphics[width=.7\linewidth]{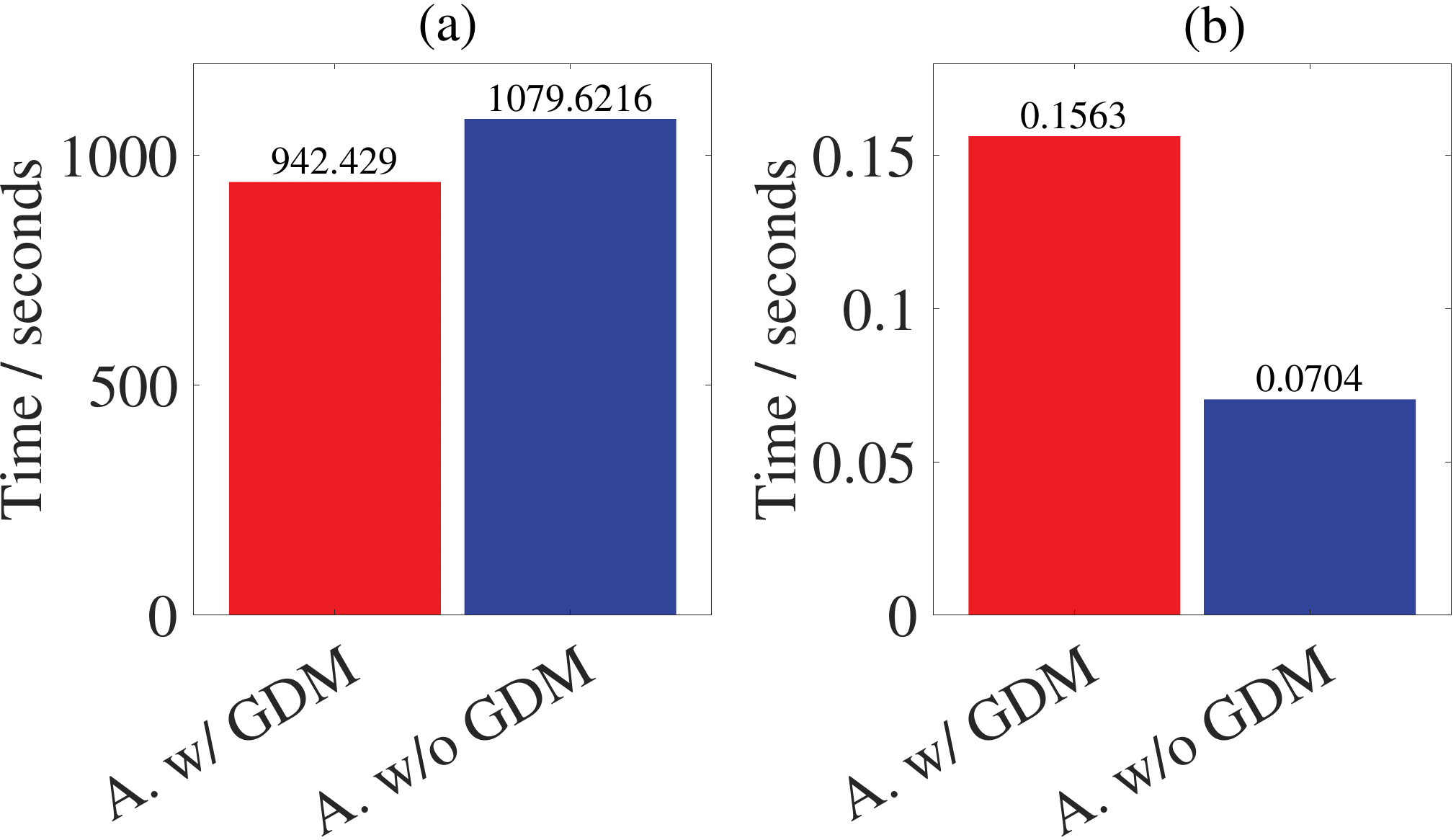}
    \caption{Comparison of (a) total GPU time for convergence achieved by GDM-based action exploration and without action exploration, and (b) GPU time per epoch for different conditions achieved by GDM-based action exploration and without action exploration.
   }  
    \label{fig:GPU Time Action exp}
\end{figure} 
\subsubsection{ \textcolor{black}{GDM-based Action Exploration}}
Fig. ~\ref{fig:sum_rate_reward_action_com} illustrates the sum rate and reward throughout the learning process, with and without action exploration. As shown in Fig. ~\ref{fig:sum_rate_reward_action_com}(a), although the sum rate achieved by A. w/ GDM in the early training stage is lower than that of A. w/o GDM, it converges more quickly to a stable optimal value. This rapid convergence indicates the superior long-term performance of A. w/ GDM. Fig.~\ref{fig:sum_rate_reward_action_com}(b) demonstrates that when the sum rate stabilizes, the MA reward of A. w/ GDM is consistently higher than that of A. w/o GDM, with the reward trend closely mirroring the sum rate. Furthermore, A. w/ GDM exhibits less reward fluctuation compared to A. w/o GDM, further confirming the robustness and stability of GDM. While both algorithms reach a plateau in the sum rate around 15,000 epochs, A. w/ GDM maintains a higher sum rate thereafter, highlighting its sustained performance advantage.

Figure \ref{fig:GPU Time Action exp} illustrates the total GPU time required for training convergence and the average GPU time per epoch. As shown in Fig. \ref{fig:GPU Time Action exp}(a), the total GPU time needed for A. w/ GDM to reach convergence is approximately 8.3$\%$ lower than that of A. w/o GDM. However, Fig. \ref{fig:GPU Time Action exp}(b) reveals that the GPU time required for each epoch in A. w/ GDM is more than twice that of A. w/o GDM. This increased per-epoch GPU time can be attributed to the additional complexity introduced by the GDM-based action exploration. Despite this higher computational time per epoch, the rapid convergence of A. w/ GDM effectively compensates for the additional time, resulting in an overall reduction in total GPU time for convergence.
This underscores the superior performance of A. w/ GDM in terms of the total GPU time required for convergence, which demonstrates the effectiveness and efficiency of incorporating GDM in the action exploration.

\subsubsection{ \textcolor{black}{GDM-based Reward Exploration}}
\begin{figure}[t]
    \centering
    \includegraphics[width=.7\linewidth]{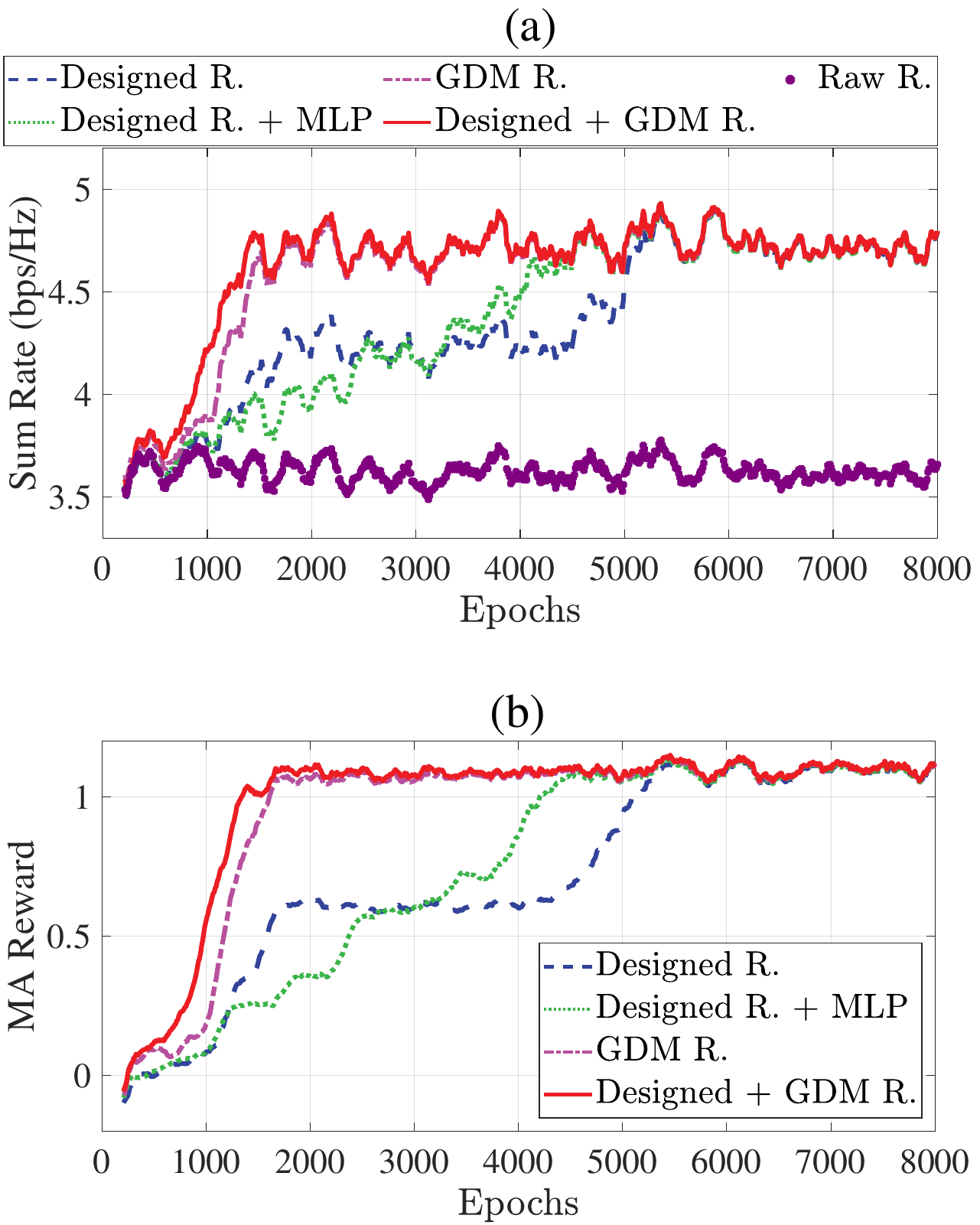}
    \caption{Comparison of (a) the sum rate and (b) MA reward for different reward design cases.
   }  
    \label{fig:Reward_sumrate}
\end{figure}

Fig.~\ref{fig:Reward_sumrate} (a) demonstrates the effectiveness of incorporating expert knowledge in reward design, with the Designed R. outperforming the Raw R. The Designed R.+MLP method further enhances this foundation by introducing exploratory capabilities through an MLP, resulting in improved performance and highlighting the potential benefits of reward exploration. Although the GDM R. converges slightly slower than the best-performing Designed+GDM R., it still exhibits significant exploratory power, surpassing the performance of the earlier methods. The Designed+GDM R. achieves the best overall performance, rapidly accelerating training convergence and attaining the highest sum rate. This underscores the effectiveness of integrating GDM into reward design for complex environments, leveraging the strengths of both expert knowledge and exploratory techniques.
\begin{figure}[t]
    \centering
    \includegraphics[width=.8\linewidth]{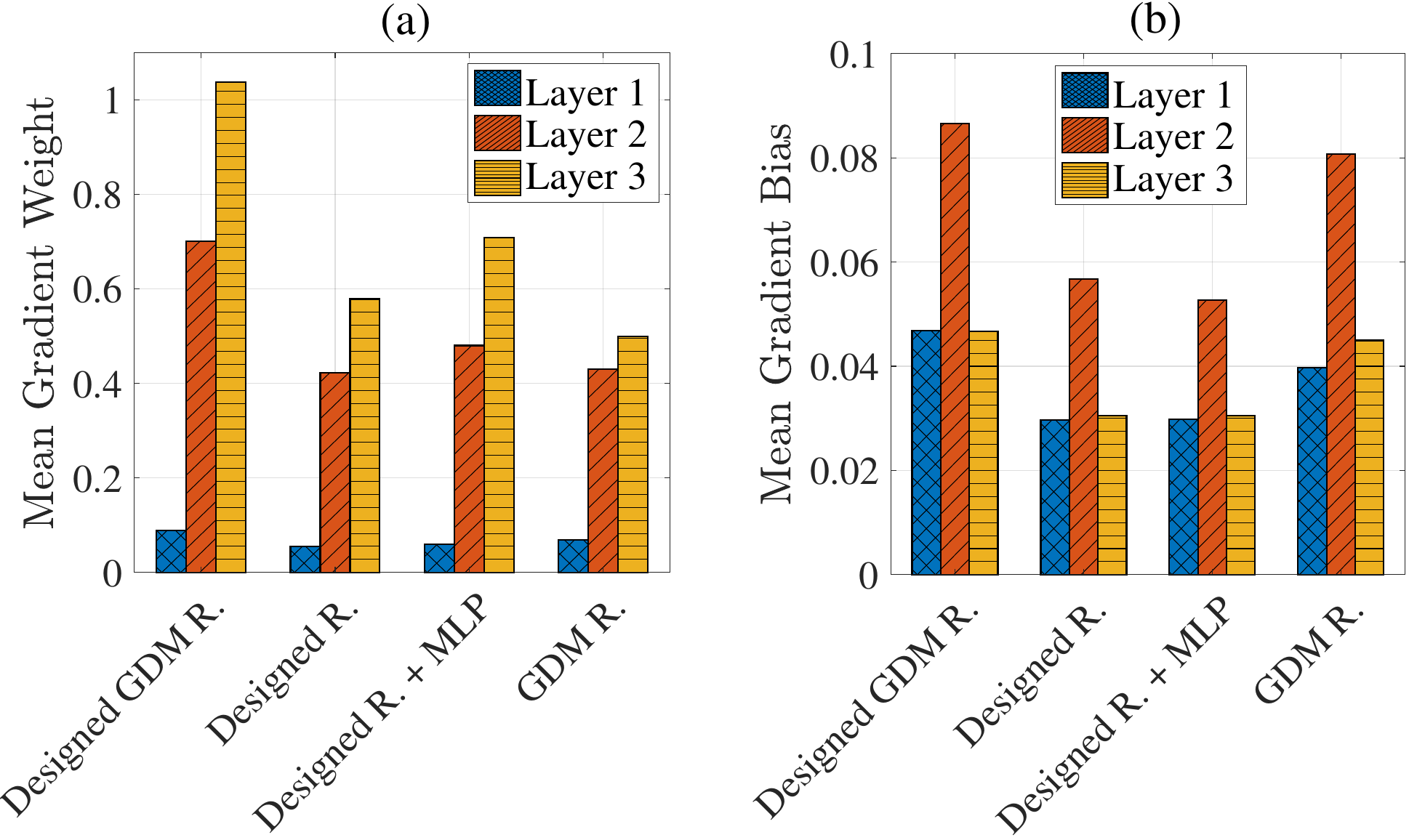}
    \caption{Comparison of (a) the mean gradient weight and (b) the mean gradient bias for different reward design cases.
   }  
    \label{fig:grad_loss}
    \vspace{-6pt}
\end{figure}
\begin{figure}[t]
    \centering
    \includegraphics[width=0.8\linewidth]{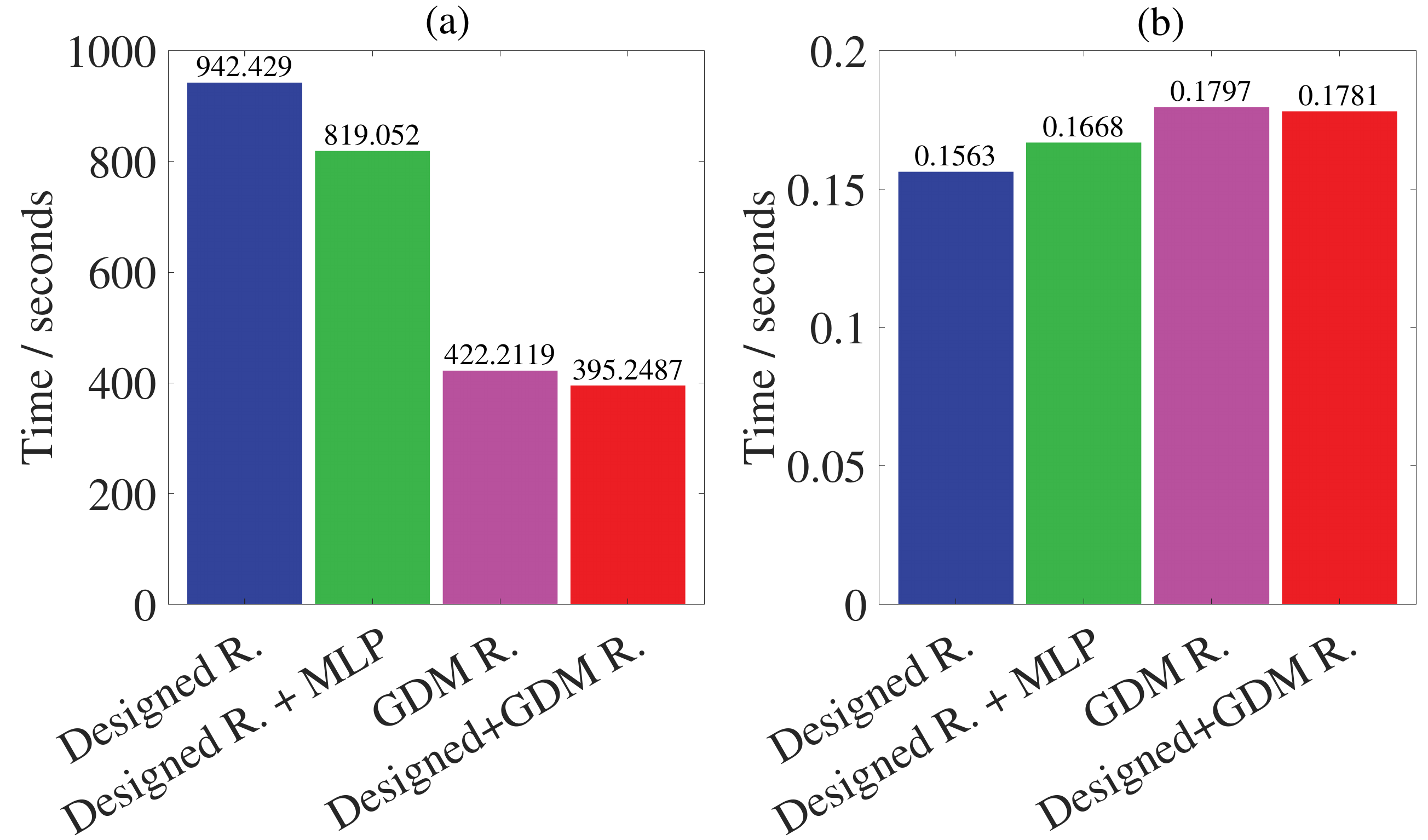}
    \caption{Comparison of (a) total GPU time for convergence and (b) GPU time per epoch for different reward design cases.
   }  
    \label{fig:GPU_time}
    \vspace{-3pt}
\end{figure}

Fig. \ref{fig:grad_loss} presents a comprehensive analysis of gradient weights and biases across different layers and reward design cases. As shown in Fig. \ref{fig:grad_loss}(a), the Designed+GDM R. approach exhibits the highest gradient weights in Layer 3, followed by Layer 2, while Layer 1 maintains minimal weights. This hierarchical distribution pattern persists across all cases with varying magnitudes. Fig. \ref{fig:grad_loss}(b) reveals gradient bias distributions ranging from 0.03 to 0.09, with Layer 2 showing notably elevated bias values (up to 0.85) in the Designed GDM R. case.
The superior performance of Designed+GDM R. can be attributed to its optimal combination of high gradient weights and well-distributed bias values across layers, leading to rapid convergence and training stability. Although the GDM R. case shows the lowest weight magnitude, it achieves relatively rapid convergence due to comparable gradient bias magnitude with Designed GDM R. Similarly, Designed R. + MLP demonstrates faster convergence than Designed R., corresponding to its higher gradient weight values while maintaining comparable gradient bias values.

Fig.~\ref{fig:GPU_time} presents the total GPU time and GPU time per epoch required for convergence across different reward design cases. The Designed+GDM R. method achieves the best overall performance, reducing the total GPU time by over 58$\%$ in comparison to the Designed R., despite a $13.9\%$ increase in per-epoch computation. This demonstrates its efficiency in balancing rapid convergence with computational cost. The GDM R. method also significantly reduces total GPU time by $55\%$ relative to Designed R., even though it requires the most time per epoch. Its faster convergence compensates for this, making it computationally efficient overall. On the other hand, Designed R.+MLP shows a little lower total GPU time than Designed R., but its per-epoch time increases by $6.7\%$, indicating that while MLP boots performance, it comes with a higher computation cost per epoch.

\subsubsection{ \textcolor{black}{GDM-based State Exploration}}

\begin{figure*}[t]
    \centering
    \includegraphics[width=.9\textwidth]{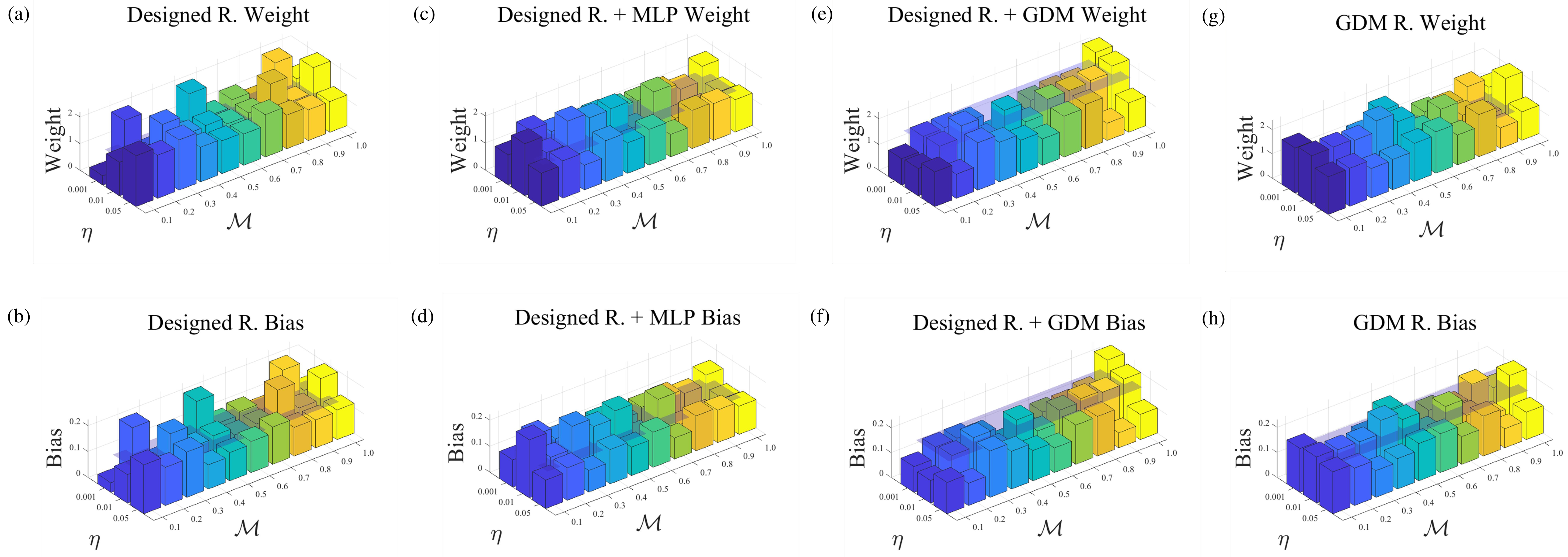}
    \caption{Comparison of gradient loss for (a) the weight sum of Designed R., (b) the bias sum of Designed R., (c) the weight sum of Designed R.+MLP, (d) the bias sum of Designed R.+MLP, (e) the weight sum of Designed R.+GDM, (f) the bias sum of Designed R.+GDM, (g) the weight sum of GDM R., and (h) the bias sum of GDM R.with different $\eta$ and $\mathcal{M}$ across different layers.
   }  
    \label{fig:State_exp_grad}
    \vspace{-6pt}
\end{figure*}

Based on our investigation, both gradient weights and biases play crucial roles in the convergence of the learning process. We conducted a comprehensive parameter sweep across two hyperparameters: i) the maximum probability $\mathcal{M}$ of the state exploration, and ii) the update rate $\eta$ at which this exploration probability increases. Fig. \ref{fig:State_exp_grad} compares the sum of gradient weights and biases across all layers under different hyperparameter configurations (the 3D bar chart) and the baseline without state exploration (the shadow plane).  
The results demonstrate that weights (range: 0-2) and biases (range: 0-0.2) exhibit distinct sensitivity patterns to these hyperparameters in our FD wireless communication network. State exploration has a dual impact on training speed. To identify the optimal configuration for each reward design case, we selected the best hyperparameter combination by comparing against the baseline. The Designed R. benefits from high $\mathcal{M}$ (0.9) with conservative $\eta$ (0.001), while the Designed R. + MLP performs best with moderate $\mathcal{M}$ (0.3) and increased $\eta$ (0.01). The Designed R. + GDM achieves optimal performance with $\mathcal{M}=1.0$ and conservative $\eta$ (0.001), and the GDM R. requires moderate $\mathcal{M}$ (0.4) with elevated $\eta$ (0.01). These carefully selected configurations ensure optimal gradient characteristics, promoting stable and efficient training dynamics across all model variants.

\begin{figure}[t]
    \centering
    \includegraphics[width=.7\linewidth]{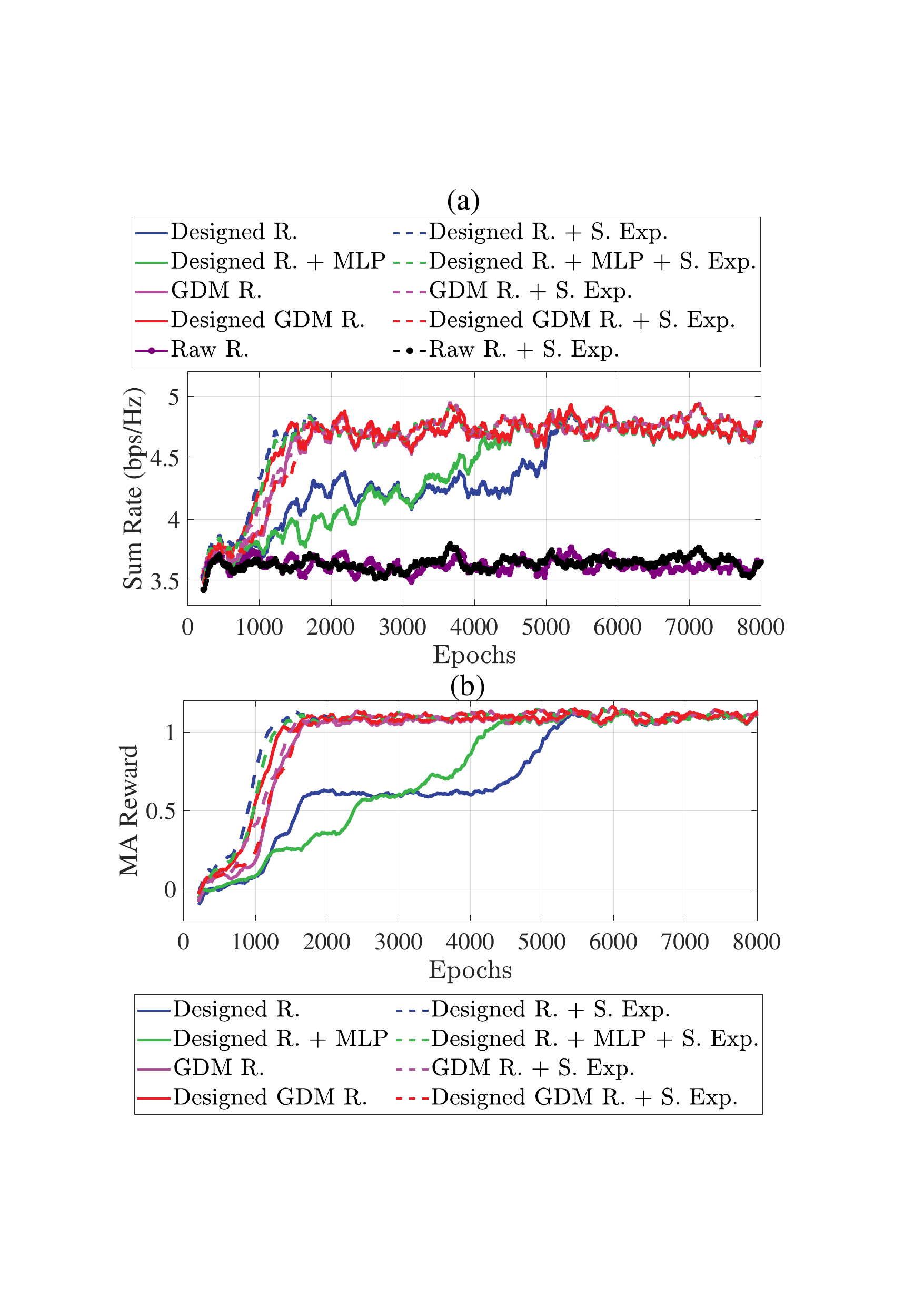}
    \caption{Comparison of (a) the sum rate and (b) MA reward for different reward design cases with GDM-based state exploration and without GDM-based state exploration.
   }  
    \label{fig:state_exp_sumrate_reward}
\end{figure}
Fig. \ref{fig:state_exp_sumrate_reward} compares sum rates and rewards across different reward design cases using optimally selected hyperparameters. The impact of state exploration varies among the cases. In the Designed R. and Designed R. + MLP cases, tuned state exploration significantly accelerates convergence from epoch 5000 to 2000, mitigating the delayed convergence caused by suboptimal reward designs. The GDM R. case shows modest improvements with state exploration. Surprisingly, the Designed GDM R. case exhibits slightly decreased learning speed when supplemented with additional state exploration, suggesting a performance ceiling where near-optimal training efficiency renders additional GDM-based state dataset replacements counterproductive. The Raw R. case remains an exception, maintaining stable but suboptimal sum rates around 3.5 regardless of state exploration.

\begin{figure}[!t]
    \centering
    \includegraphics[width=.8\linewidth]{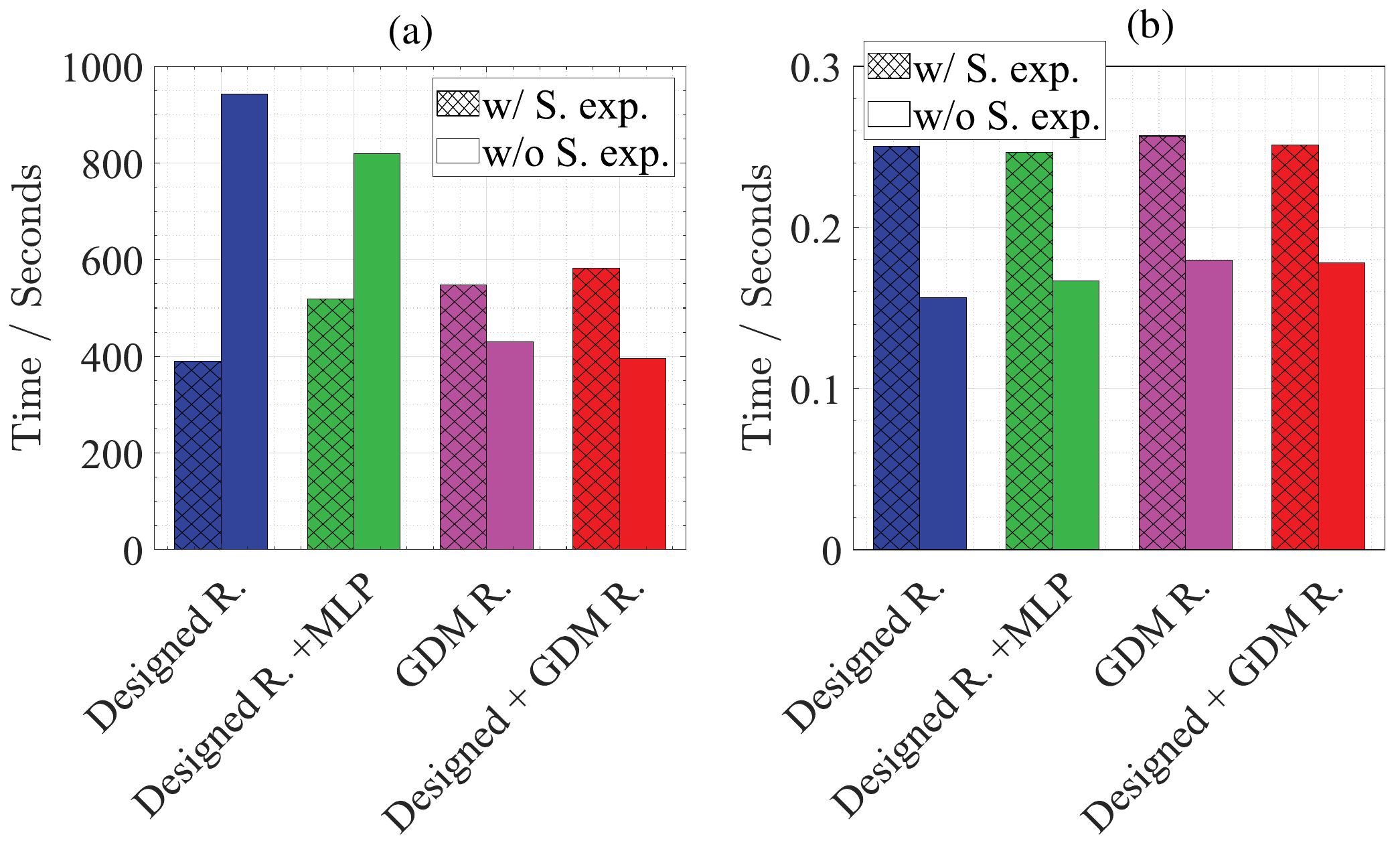}
    \caption{Comparison of (a) total GPU time for convergence and (b) GPU time per epoch for different reward design cases with GDM-based state exploration and without GDM-based state exploration.
   }  
    \label{fig:state_exp_GPU_time}
\end{figure} 
Fig.~\ref{fig:state_exp_GPU_time} compares the total and per-epoch GPU time required for convergence across different reward design methods, with and without state exploration. While state exploration increases per-epoch computational overhead across all methods, it demonstrates potential for reducing total GPU time through enhanced training efficiency.
The Designed R. and Designed R.+MLP methods show significant reductions in total GPU time with state exploration, indicating that improved state space exploration expedites convergence to optimal solutions. Conversely, Designed+GDM R. and GDM R. experience increased total GPU time with state exploration, supporting the hypothesis that near-optimal reward designs may not benefit from additional state dataset replacements. The consistent increase in per-epoch computational costs across all methods reflects the additional overhead of complex state sampling strategies.
\vspace{-8pt}
\section{Conclusions}
\label{sec:conclusion}
 
Addressing training inefficiency is critical for enabling real-world DRL deployment in dynamic wireless networks, where periodic retraining under time-varying environment risks prohibitive computational delays and energy costs. This paper presented the D2RL framework to address training inefficiencies for resource allocation in dynamic wireless networks by integrating GDMs across three core DRL components. Experimental results demonstrated that GDM-driven reward exploration, particularly when combined with expert-designed reward structures, achieved the fastest convergence and highest sum-rate performance by balancing exploration with domain-specific knowledge. While GDM-enhanced action exploration reduced total training time despite increased per-epoch computational costs, state-space diversification via GDMs exhibited dual effects: it accelerated convergence in scenarios with suboptimal reward designs but hindered performance when reward functions were already optimized, underscoring the necessity of context-aware hyperparameter tuning for state replacement probability and update rates. The framework’s sensitivity analysis revealed critical trade-offs between computational overhead and training efficiency, with optimally tuned state exploration complementing reward refinement to mitigate inefficiencies in complex channel environments. These findings advocate for systematic co-design of DRL components, providing a pathway toward adaptive resource allocation in real-world mobile networks. 
 
\bibliographystyle{IEEEtran}
\bibliography{IEEEabrv,mybib}

\begin{thebibliography}{10}
\providecommand{\url}[1]{#1}
\csname url@samestyle\endcsname
\providecommand{\newblock}{\relax}
\providecommand{\bibinfo}[2]{#2}
\providecommand{\BIBentrySTDinterwordspacing}{\spaceskip=0pt\relax}
\providecommand{\BIBentryALTinterwordstretchfactor}{4}
\providecommand{\BIBentryALTinterwordspacing}{\spaceskip=\fontdimen2\font plus
\BIBentryALTinterwordstretchfactor\fontdimen3\font minus
  \fontdimen4\font\relax}
\providecommand{\BIBforeignlanguage}[2]{{%
\expandafter\ifx\csname l@#1\endcsname\relax
\typeout{** WARNING: IEEEtran.bst: No hyphenation pattern has been}%
\typeout{** loaded for the language `#1'. Using the pattern for}%
\typeout{** the default language instead.}%
\else
\language=\csname l@#1\endcsname
\fi
#2}}
\providecommand{\BIBdecl}{\relax}
\BIBdecl

\bibitem{hieu2021optimal}
N.~Q. Hieu, D.~T. Hoang, D.~Niyato, and D.~I. Kim, ``Optimal power allocation
  for rate splitting communications with deep reinforcement learning,''
  \emph{IEEE Wireless Commun. Lett.}, vol.~10, no.~12, pp. 2820--2823, 2021.

\bibitem{gao2024uav}
Y.~Gao, X.~Yuan, D.~Yang, Y.~Hu, Y.~Cao, and A.~Schmeink, ``{UAV}-assisted
  {MEC} system with mobile ground terminals: {DRL}-based joint terminal
  scheduling and {UAV} 3{D} trajectory design,'' \emph{{IEEE} Trans. Veh.
  Technol.}, vol.~73, no.~7, pp. 10\,164--10\,180, 2024.

\bibitem{he2023full}
Z.~He, W.~Xu, H.~Shen \emph{et~al.}, ``Full-duplex communication for {ISAC}:
  Joint beamforming and power optimization,'' \emph{IEEE J. Sel. Areas
  Commun.}, vol.~41, no.~9, pp. 2920--2936, 2023.

\bibitem{chauhan2024full}
A.~Chauhan, A.~Jaiswal, and C.~Tellambura, ``Full-duplex cooperative {NOMA}
  with signal space diversity: Minimizing {SIC} operations,'' \emph{IEEE
  Wireless Commun. Lett.}, 2024, doi: {10.1109/LWC.2024.3429236}.

\bibitem{nayak2024drl}
N.~Nayak, S.~Kalyani, and H.~A. Suraweera, ``A {DRL} approach for
  {RIS}-assisted full-duplex {UL} and {DL} transmission: Beamforming, phase
  shift and power optimization,'' \emph{IEEE Trans. Wireless Commun.}, vol.~23,
  no.~10, pp. 14\,652--14\,666, 2024.

\bibitem{kumar2020conservative}
A.~Kumar, A.~Zhou, G.~Tucker, and S.~Levine, ``Conservative q-learning for
  offline reinforcement learning,'' \emph{Adv. Neural Inf. Process. Syst.},
  vol.~33, pp. 1179--1191, 2020.

\bibitem{agarwal2020optimistic}
R.~Agarwal, D.~Schuurmans, and M.~Norouzi, ``An optimistic perspective on
  offline reinforcement learning,'' in \emph{International Conference on
  Machine Learning}.\hskip 1em plus 0.5em minus 0.4em\relax PMLR, 2020, pp.
  104--114.

\bibitem{muhammad2021deep}
G.~Muhammad and M.~S. Hossain, ``Deep-reinforcement-learning-based sustainable
  energy distribution for wireless communication,'' \emph{{IEEE} Wireless
  Commun.}, vol.~28, no.~6, pp. 42--48, 2021.

\bibitem{jin2022vwp}
Y.-L. Jin, Z.-Y. Ji, D.~Zeng, and X.-P. Zhang, ``{VWP}: An efficient
  {DRL}-based autonomous driving model,'' \emph{IEEE Trans. Multimedia},
  vol.~26, pp. 2096--2108, 2022.

\bibitem{qian2024offline}
T.~Qian, Z.~Liang, C.~Shao, H.~Zhang, Q.~Hu, and Z.~Wu, ``Offline {DRL} for
  price-based demand response: Learning from suboptimal data and beyond,''
  \emph{IEEE Trans. Smart Grid}, vol.~15, no.~5, pp. 4618--4635, 2024.

\bibitem{wang2021provably}
L.~Wang, Z.~Yang, and Z.~Wang, ``Provably efficient causal reinforcement
  learning with confounded observational data,'' \emph{Adv. Neural Inf.
  Process. Syst.}, vol.~34, pp. 21\,164--21\,175, 2021.

\bibitem{wang2022vrl3}
C.~Wang, X.~Luo, K.~Ross, and D.~Li, ``Vrl3: A data-driven framework for visual
  deep reinforcement learning,'' \emph{Adv. Neural Inf. Process. Syst.},
  vol.~35, pp. 32\,974--32\,988, 2022.

\bibitem{mohi2024optimizing}
N.~Mohi Ud~Din, A.~Assad, S.~Ul~Sabha, and M.~Rasool, ``Optimizing deep
  reinforcement learning in data-scarce domains: A cross-domain evaluation of
  double {DQN} and dueling {DQN},'' \emph{International Journal of System
  Assurance Engineering and Management}, pp. 1--12, 2024.

\bibitem{chen2023structure}
J.~Chen, W.~Liu, D.~E. Quevedo, S.~R. Khosravirad, Y.~Li, and B.~Vucetic,
  ``Structure-enhanced {DRL} for optimal transmission scheduling,''
  \emph{{IEEE} Trans. Wireless Commun.}, vol.~23, no.~1, pp. 379--393, 2023.

\bibitem{michaud2020understanding}
E.~J. Michaud, A.~Gleave, and S.~Russell, ``Understanding learned reward
  functions,'' \emph{arXiv preprint arXiv:2012.05862}, 2020.

\bibitem{lyu2020movement}
Z.~Lyu, C.~Ren, and L.~Qiu, ``Movement and communication co-design in
  multi-{UAV} enabled wireless systems via {DRL},'' in \emph{2020 IEEE 6th
  International Conference on Computer and Communications (ICCC)}.\hskip 1em
  plus 0.5em minus 0.4em\relax IEEE, 2020, pp. 220--226.

\bibitem{du2024enhancing}
H.~Du, R.~Zhang, Y.~Liu, J.~Wang, Y.~Lin, Z.~Li, D.~Niyato, J.~Kang, Z.~Xiong,
  S.~Cui \emph{et~al.}, ``Enhancing deep reinforcement learning: A tutorial on
  generative diffusion models in network optimization,'' \emph{{IEEE} Commun.
  Surveys Tutorials}, vol.~26, no.~4, pp. 2611--2646, 2024.

\bibitem{sohl2015deep}
J.~Sohl-Dickstein, E.~Weiss, N.~Maheswaranathan, and S.~Ganguli, ``Deep
  unsupervised learning using nonequilibrium thermodynamics,'' in
  \emph{International Conference on Machine Learning}.\hskip 1em plus 0.5em
  minus 0.4em\relax PMLR, 2015, pp. 2256--2265.

\bibitem{zhi2022deep}
Y.~Zhi, J.~Tian, X.~Deng, J.~Qiao, and D.~Lu, ``Deep reinforcement
  learning-based resource allocation for {D2D} communications in heterogeneous
  cellular networks,'' \emph{Digit. Commun. Netw.}, vol.~8, no.~5, pp.
  834--842, 2022.

\bibitem{luong2021deep}
P.~Luong, F.~Gagnon, L.-N. Tran, and F.~Labeau, ``Deep reinforcement
  learning-based resource allocation in cooperative {UAV}-assisted wireless
  networks,'' \emph{{IEEE} Trans. Wireless Commun.}, vol.~20, no.~11, pp.
  7610--7625, 2021.

\bibitem{wang2023hybrid}
L.~Wang, W.~Wu, F.~Zhou, Q.~Wu, O.~A. Dobre, and T.~Q. Quek, ``Hybrid
  hierarchical {DRL} enabled resource allocation for secure transmission in
  multi-{IRS}-assisted sensing-enhanced spectrum sharing networks,''
  \emph{{IEEE} Trans. Wireless Commun.}, vol.~23, no.~6, pp. 6330--6346, 2024.

\bibitem{wang2020drl}
X.~Wang, Y.~Zhang, R.~Shen, Y.~Xu, and F.-C. Zheng, ``{DRL}-based
  energy-efficient resource allocation frameworks for uplink {NOMA} systems,''
  \emph{{IEEE} Internet Things J.}, vol.~7, no.~8, pp. 7279--7294, 2020.

\bibitem{zhang2023drl}
H.~Zhang, H.~Wang, Y.~Li, K.~Long, and A.~Nallanathan, ``{DRL}-driven dynamic
  resource allocation for task-oriented semantic communication,'' \emph{{IEEE}
  Trans. Commun.}, vol.~71, no.~7, pp. 3992--4004, 2023.

\bibitem{tran2023multi}
D.-D. Tran, S.~K. Sharma, V.~N. Ha, S.~Chatzinotas, and I.~Woungang,
  ``Multi-agent {DRL} approach for energy-efficient resource allocation in
  {URLLC}-enabled grant-free {NOMA} systems,'' \emph{IEEE Open J. Commun.
  Soc.}, vol.~4, pp. 1470--1486, 2023.

\bibitem{janner2022planning}
M.~Janner, Y.~Du, J.~B. Tenenbaum, and S.~Levine, ``Planning with diffusion for
  flexible behavior synthesis,'' \emph{arXiv preprint arXiv:2205.09991}, 2022.

\bibitem{kang2024hybrid}
Y.~Kang, J.~Wen, J.~Kang, T.~Zhang, H.~Du, D.~Niyato, R.~Yu, and S.~Xie,
  ``Hybrid-generative diffusion models for attack-oriented twin migration in
  vehicular metaverses,'' \emph{arXiv preprint arXiv:2407.11036}, 2024.

\bibitem{du2024diffusion}
H.~Du, Z.~Li, D.~Niyato, J.~Kang, Z.~Xiong, H.~Huang, and S.~Mao,
  ``Diffusion-based reinforcement learning for edge-enabled {AI}-generated
  content services,'' \emph{{IEEE} Trans. Mobile Comput.}, vol.~23, no.~9, pp.
  8902--8918, 2024.

\bibitem{qin2024diffusiongpt}
J.~Qin, J.~Wu, W.~Chen, Y.~Ren, H.~Li, H.~Wu, X.~Xiao, R.~Wang, and S.~Wen,
  ``Diffusiongpt: {LLM}-driven text-to-image generation system,'' \emph{arXiv
  preprint arXiv:2401.10061}, 2024.

\bibitem{seid2021multi}
A.~M. Seid, G.~O. Boateng, B.~Mareri, G.~Sun, and W.~Jiang, ``Multi-agent {DRL}
  for task offloading and resource allocation in multi-{UAV} enabled iot edge
  network,'' \emph{{IEEE} Trans. Netw. Service Manag.}, vol.~18, no.~4, pp.
  4531--4547, 2021.

\bibitem{zhang2022drl}
S.~Zhang, H.~Gu, K.~Chi, L.~Huang, K.~Yu, and S.~Mumtaz, ``{DRL}-based partial
  offloading for maximizing sum computation rate of wireless powered mobile
  edge computing network,'' \emph{{IEEE} Trans. Wireless Commun.}, vol.~21,
  no.~12, pp. 10\,934--10\,948, 2022.

\bibitem{hao2022delay}
Y.~Hao, F.~Li, C.~Zhao, and S.~Yang, ``Delay-oriented scheduling in {5G}
  downlink wireless networks based on reinforcement learning with partial
  observations,'' \emph{{IEEE} ACM Trans. Netw.}, vol.~31, no.~1, pp. 380--394,
  2022.

\bibitem{kasi2022d}
S.~K. Kasi, U.~S. Hashmi, S.~Ekin, A.~Abu-Dayya, and A.~Imran, ``{D-RAN}: A
  {DRL}-based demand-driven elastic user-centric {RAN} optimization for {6G} \&
  beyond,'' \emph{{IEEE} Trans. Cogn. Commun. Netw.}, vol.~9, no.~1, pp.
  130--145, 2022.

\bibitem{swistak2024qos}
E.~Swistak, M.~Roshdi, R.~German, and M.~Harounabadi, ``{QoS-DRAMA}: Quality of
  service aware drl-based adaptive mid-level resource allocation scheme,'' in
  \emph{IEEE INFOCOM 2024-IEEE Conference on Computer Communications Workshops
  (INFOCOM WKSHPS)}.\hskip 1em plus 0.5em minus 0.4em\relax IEEE, 2024, pp.
  1--6.

\bibitem{al2022self}
Y.~Al-Eryani and E.~Hossain, ``Self-organizing mmwave {MIMO} cell-free networks
  with hybrid beamforming: A hierarchical {DRL}-based design,'' \emph{{IEEE}
  Trans. Commun.}, vol.~70, no.~5, pp. 3169--3185, 2022.

\bibitem{he2023reinforcement}
J.~He, B.~Tan, and Y.~Gao, ``Reinforcement learning based downlink {OFDMA}
  scheduling for time-sensitive wifi networks,'' in \emph{2023 IEEE/CIC
  International Conference on Communications in China (ICCC)}.\hskip 1em plus
  0.5em minus 0.4em\relax IEEE, 2023, pp. 1--6.

\bibitem{chai2024drl}
R.~Chai, G.~Yang, L.~Liu, and Q.~Chen, ``{DRL}-based dynamic resource
  allocation for multi-beam satellite systems,'' \emph{{IEEE} Trans. Netw.
  Serv. Manag.}, vol.~21, no.~4, pp. 3829--3845, 2024.

\bibitem{song2020denoising}
J.~Song, C.~Meng, and S.~Ermon, ``Denoising diffusion implicit models,''
  \emph{arXiv preprint arXiv:2010.02502}, 2020.

\bibitem{yu2024attention}
J.~Yu, A.~Alhilal, T.~Zhou, P.~Hui, and D.~H. Tsang, ``Attention-based
  {QoE}-aware digital twin empowered edge computing for immersive virtual
  reality,'' \emph{{IEEE} Trans. Wireless Commun.}, vol.~23, no.~9, pp.
  11\,276--11\,290, 2024.

\bibitem{hu2023intelligent}
H.~Hu, D.~Wu, F.~Zhou, X.~Zhu, R.~Q. Hu, and H.~Zhu, ``Intelligent resource
  allocation for edge-cloud collaborative networks: A hybrid {DDPG-D3QN}
  approach,'' \emph{{IEEE} Trans. Veh. Technol.}, vol.~72, no.~8, pp.
  10\,696--10\,709, 2023.

\bibitem{miao2024utility}
J.~Miao, S.~Bai, S.~Mumtaz, Q.~Zhang, and J.~Mu, ``Utility-oriented
  optimization for video streaming in {UAV}-aided {MEC} network: A {DRL}
  approach,'' \emph{{IEEE} Trans. Green Commun. Netw.}, vol.~8, no.~2, pp.
  878--889, 2024.

\bibitem{zhou2024federated}
T.~Zhou, J.~Yu, J.~Zhang, and D.~H. Tsang, ``Federated prompt-based decision
  transformer for customized {VR} services in mobile edge computing system,''
  \emph{arXiv preprint arXiv:2402.09729}, 2024.

\end{thebibliography}

\end{document}